\documentclass[letterpaper]{article} % DO NOT CHANGE THIS
\usepackage{aaai25}  % DO NOT CHANGE THIS
\usepackage{times}  % DO NOT CHANGE THIS
\usepackage{helvet}  % DO NOT CHANGE THIS
\usepackage{courier}  % DO NOT CHANGE THIS
\usepackage[hyphens]{url}  % DO NOT CHANGE THIS
\usepackage{graphicx} % DO NOT CHANGE THIS
\usepackage{natbib}  % DO NOT CHANGE THIS AND DO NOT ADD ANY OPTIONS TO IT
\usepackage{caption} % DO NOT CHANGE THIS AND DO NOT ADD ANY OPTIONS TO IT
\usepackage{algorithm}
\usepackage{algorithmic}
\usepackage{amssymb}
\usepackage{booktabs}
\usepackage{multirow}
\usepackage{algorithm}
\usepackage{algorithmic}
\usepackage{amssymb}
\usepackage{booktabs}
\usepackage{multirow}
\usepackage{url}
\usepackage{amsmath}
\usepackage{amsthm}
\usepackage{amsfonts}
\usepackage{newfloat}
\usepackage{listings}

\DeclareCaptionStyle{ruled}{labelfont=normalfont,labelsep=colon,strut=off} % DO NOT CHANGE THIS
\floatstyle{ruled}
\newfloat{listing}{tb}{lst}{}
\floatname{listing}{Listing}
\title{Enhancing Masked Time-Series Modeling via Dropping Patches}
\author{
    Tianyu Qiu\textsuperscript{\rm 1}, 
    Yi Xie\textsuperscript{\rm 1}\thanks{Corresponding author.}, 
    Yun Xiong\textsuperscript{\rm 1}, 
    Hao Niu\textsuperscript{\rm 1}, 
    Xiaofeng Gao\textsuperscript{\rm 2}
}
\affiliations{
    \textsuperscript{\rm 1}Shanghai Key Lab of Data Science, School of Computer Science, Fudan University, Shanghai, China\\
    \{tyqiu22, yixie18, yunx, hniu18\}@fudan.edu.cn\\
    \textsuperscript{\rm 2}MoE Key Lab of Artificial Intelligence, Shanghai Jiao Tong University, Shanghai, China\\
    gao-xf@cs.sjtu.edu.cn
}

\usepackage{bibentry}

\newtheorem{lemma}{Lemma}
\newtheorem{corollary}{Corollary}
\begin{document}

\maketitle

\begin{abstract}
This paper explores how to enhance existing masked time-series modeling by randomly dropping sub-sequence level patches of time series. On this basis, a simple yet effective method named \textbf{DropPatch} is proposed, which has two remarkable advantages: 1) It improves the pre-training efficiency by a square-level advantage; 2) It provides additional advantages for modeling in scenarios such as in-domain, cross-domain, few-shot learning and cold start. This paper conducts comprehensive experiments to verify the effectiveness of the method and analyze its internal mechanism. Empirically, DropPatch strengthens the attention mechanism, reduces information redundancy and serves as an efficient means of data augmentation. Theoretically, it is proved that DropPatch slows down the rate at which the Transformer representations collapse into the rank-1 linear subspace by randomly dropping patches, thus optimizing the quality of the learned representations. 
\end{abstract}


\section{Introduction}
In recent years, masked modeling has emerged as a prevalent self-supervised method in various fields, including natural language processing \cite{BERT,ROBERTa} and computer vision \cite{Data2VEC,MaskedAE,BeiT}. This technique improves representation learning by reconstructing masked content based on unmasked parts. Masked modeling has also been adapted for time-series analysis. A notable advancement involves segmenting time-series into patches (sub-sequence) and applying a patch-level masking strategy, which has received considerable attention since its inception \cite{PatchTST}. This method not only shows promising performance in transfer learning, but also significantly enhances supervised forecasting by employing self-supervised pre-training to initialize model parameters, consistent with recent findings \cite{nevertrainfrom}. 
Building upon the patching technique, numerous time-series foundation model works have emerged and achieve significant performance in time-series forecasting \cite{moment,moirai}.

\begin{figure*}[t]
  \centering
  \includegraphics[width=\textwidth]{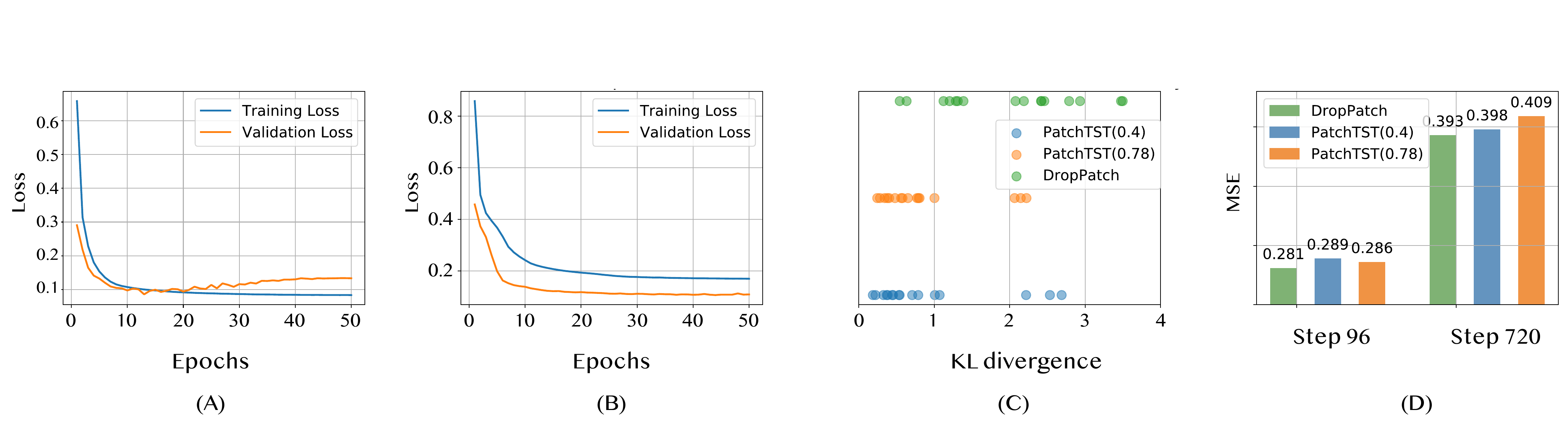}
  \caption{(A) The loss curve of PatchTST with lower mask ratio 0.4 (official implementation); (B) The loss curve of DropPatch (unless otherwise stated, the drop ratio and mask ratio is 0.6 and 0.4 throughout this paper); (C) The Kullback-Leibler (KL) divergence between the attention coefficients of the final encoder layer and a uniform distribution, where each dot represents an individual attention head. A larger KL divergence indicates that this set of attention distributions is farther from a uniform distribution and thus more focused. PatchTST(0.78) refers to the PatchTST configured with a mask ratio of 0.78, matching the number of visible patches in DropPatch. (D) Comparison of MSE metrics between PatchTST and DropPatch with forecasting steps $T \in \{96, 720\}$ on ETTm1.}
  \label{fig:intro}
\end{figure*}

Despite its potential, we observed that masked time-series modeling, represented by PatchTST \cite{PatchTST}, faces a dilemma.
A relatively low mask ratio reduces effectiveness in learning useful features \cite{ MaskedAE, maskTheo}. Given the characteristic of periodicity and repetitive pattern of time-series data, the masked patch can be recovered with little high-level understanding of the underlying patterns, leading to superficial learning and over-fitting as shown in Figure \ref{fig:intro} (A).
A natural idea is to increase the mask ratio, but another issue emerges: the presence of an excessive number of masked patches can further dilute the attention mechanism's capacity to concentrate on the relevant and informative parts of data, termed as scattered attention as shown in Figure \ref{fig:intro} (C). It can lead to the degradation of downstream task performance as the representations gradually lose their distinctiveness \cite{AttentionColl1, AttentionColl2, AttentionColl3}. 

We introduce a simple yet effective strategy, \textbf{DropPatch}, to encourage learning useful features and improve the overall performance. Building on foundational time-series pre-training techniques \cite{PatchTST}, DropPatch randomly removes a predefined proportion of patches. The remaining patches are subsequently processed for masking and reconstruction. It is crucial to distinguish between dropping and masking in the context of pre-training. For a given time-series sample, the dropping operation is applied prior to masking and reconstruction. Removed patches are entirely excluded from all training processes during the current epoch. In contrast, masked patches, represented as zero tensors overlaid with positional encoding, remain part of the training process throughout the epoch.

In our empirical study, DropPatch demonstrates clear advantages in mitigating over-fitting (Figure \ref{fig:intro} (B)), enhancing attention focus (Figure \ref{fig:intro} (C)), and improving forecasting performance (Figure \ref{fig:intro} (D)). The reduction in the number of patches due to the dropping operation leads to significant improvements in computational efficiency and reduced memory consumption.

Extensive experiments validate the effectiveness of DropPatch. Through detailed experimental analysis, we uncover the underlying mechanisms driving these improvements. The DropPatch strategy enhances the attention mechanism by enabling a sharper focus on multi-scale and diverse information. It strengthens the model's ability to capture critical patterns while reducing redundancy in representation. Furthermore, our theoretical findings indicate that the random dropping of patches effectively slows the convergence of the Transformer's representations toward a rank-1 linear subspace, thereby promoting the feature diversity.

Overall, our contributions can be summarized as follows:
\begin{itemize}
\item We introduce DropPatch, a simple yet effective strategy that enhances masked time-series modeling.
\item Extensive experiments demonstrate that the DropPatch strategy improves pre-training efficiency and delivers substantial performance gains across diverse downstream tasks. Additionally, we compile comprehensive synthesized datasets to evaluate its role as a core component in foundational models for time-series analysis.
\item Through rigorous empirical and theoretical analysis, we validate the effectiveness of DropPatch and provide insights into the mechanisms driving these improvements.
%, and improving representation abilities.
\end{itemize}

\section{Method}
\begin{figure*}[ht]
  \centering
\includegraphics[width=\textwidth]{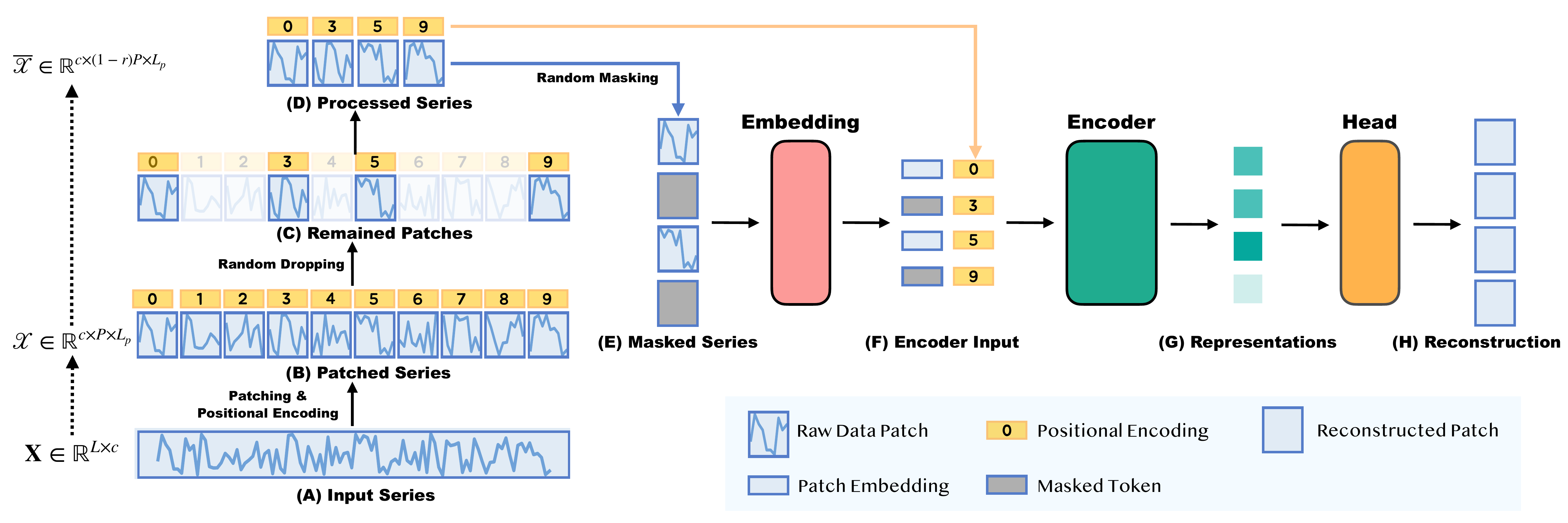}
  \caption{The overall pre-training framework of DropPatch.}
  \label{fig:structure2}
\end{figure*}

In this section, we describe the details of our proposed pre-training method, DropPatch, as shown in Fig. \ref{fig:structure2}. We denote that DropPatch is an effective pre-training strategy, the model does not perform the dropping operation during the fine-tuning stage.

% \section{Methodology}

\subsection{Patching and Channel-Independence}
For each sample of multivariate time-series $\mathbf{X}\in{\mathbb{R}^{L\times{c}}}$,
where $L$ represents the length of time-series, and 
$c$ denotes the number of channels (variates). We first split the entire time-series sample into non-overlapping subseries-level patches, which are served as input tokens to Transformer, like PatchTST \cite{PatchTST}. We permute the original data of time-series into $\mathcal{X}\in{\mathbb{R}^{c\times{P\times{L_{P}}}}}$, where $L_{P}$ denotes the length of each subseries-level patch, and $P$ denotes the total number of patches. 
% Note that two consecutive patches are non-overlapping. Here, we denote the non-overlapping content as $S$, thus $P=\lfloor \frac{(L-L_{P})}{S} \rfloor+1$.

\subsection{Dropping Patches}
After the patching operation, we will first conduct the positional encoding for these patches.
% , which is common in Transformer \cite{Transformer}:
% \begin{equation}
% \left\{
% \begin{array}{l}
% PE(pos, 2i) = \sin\left(\frac{pos}{10000^{2i/d_{\text{model}}}}\right) \\
% PE(pos, 2i+1) = \cos\left(\frac{pos}{10000^{2i/d_{\text{model}}}}\right)
% \end{array}
% \right.
% \end{equation}
% where
% $pos$ is the position of the token in the sequence, 
% $i$ is the dimension index, and 
% $d_{\text{model}}$ is the dimensionality of the token embedding. 
% In this context, the term "token" refers to each patch of values. 
The positional encoding process is designed to preserve the positional information during the self-attention computation and following the dropping operation. 
It should be noted that the positional encoding of each token is computed prior to dropping operation, ensuring that the original sequence position of each token is maintained after the removal.

We randomly drop patches in the patched time-series, which is the core idea of our proposed DropPatch. 
Let $r$ denotes the ratio of dropping with condition $0\leq r\leq 1$, implying that only $(1-r)P$ patches remains for further training and others will be directly absent in the subsequent operations. Formally, the remained patches and positional encoding will be denoted as $\bar{\mathcal{X}}\in{\mathbb{R}^{c\times{(1-r)P\times{L_{P}}}}}$, ${\overline{PE} \in {\mathbb{R}^{c\times{(1-r)P\times{d_{model}}}}}}$. 

\subsection{Representation Learning}
Subsequently, a patch-level random masking strategy is applied to generate masked data, the resultant masked dataset can be expressed as $\bar{\mathcal{X}}_{masked}\in{\mathbb{R}^{c\times{(1-r)P\times{L_{P}}}}}$. 
Given a mask ratio $m\in \left [ 1, 0 \right ] $, we denote that the number of masked patches is $\left (  1-r\right )mP$. 

The masked data is then embedded, and the previously dropped positional encodings are added back to these embeddings to formulate the encoder input $\textbf{E}$. After the encoder, we can obtain the representation $\textbf{Z}$ of the input series which can be formalized as:
\begin{equation}
    \textbf{E} = Embed(\bar{\mathcal{X}}_{masked}) + {\overline{PE}}, 
\end{equation}
\begin{equation}
    \textbf{Z} = Encoder(\textbf{E}),
\end{equation}
where $\textbf{E}, \textbf{Z} \in {\mathbb{R}^{c\times{(1-r)P\times{d_{model}}}}}$.
Finally, the representation $\textbf{Z}$ is fed into a reconstruction head to obtain the reconstruction results $\hat{X} \in {\mathbb{R}^{c\times{(1-r)P\times{L_{P}}}}}$. In the implementation, we simply adopt a linear layer as the head. We choose to use the Mean Squared Error (MSE) loss to mesure the reconstruction and the ground truth. Only the reconstructions on the masked patches are considered in the loss.

% \subsection{Theoritical Analysis}
Here, we present a corollary to describe from the perspective of representation space why DropPatch is effective, which will be validated through both experimental and theoretical approaches in the following text.
\begin{lemma}
Let $\mathrm{SAN}$ denote a self-attention layer, and consider stacking $L$ such layers. Then, under certain conditions, the representations within the stacked self-attention layers will converge to a rank-1 matrix as $L \to \infty$.
\end{lemma}

\begin{corollary}
The DropPatch strategy effectively slows down the rate at which the representation matrix of a Transformer degenerates into a rank-1 matrix.
\end{corollary}

\section{Experiments}
We perform time-series forecasting task under in-domain, cross-domain, few-shot and cold start settings to demonstrate the effectiveness of our proposed method. Furthermore, we do evaluations on two merged synthesized time-series datasets containing over 3.76 millons and 36 million data points, respectively. It worth noting that we maintain consistent drop ratio and mask ratio to be fixed across various tasks and datasets, demonstrating the effectiveness and robustness of our approach. 
% In fact, ablation experiments indicate that carefully tuning the drop ratio for specific datasets can yield improved results.

\subsubsection{Datasets} We evaluate performance of our proposed method DropPatch\footnote{In this section, we refer to DropPatch as DropPatch strategy implemented on top of the PatchTST backbone} on 12 popular datasets. For in-domain, cross-domain and few-shot experiments, Weather, ECL, Traffic and 4 ETT datasets (ETTh1, ETTh2, ETTm1, ETTm2) are inclued. In addition, we incorporate Exchange and PEMS dataset for cold start scenario in cross-domain transfer learning. 
All datasets are available on \cite{autoformer} \cite{scinet}. 
Moreover, we compile two synthesized datasets to conduct multi-dataset pre-training \cite{moment}, demonstrating the potential of DropPatch strategy in time-series foundation model.

\subsubsection{Implementation} We choose seven competitive self-supervised baseline methods, including the masked modeling method: PatchTST \cite{PatchTST}, SimMTM \cite{SimMTM}, Ti-MAE \cite{timemae}, TST \cite{TST}, the contrastive learning methods: LaST \cite{last}, CoST \cite{cost}, TS2Vec \cite{ts2vec}. We also include supervised methods iTransoformer \cite{itransformer}, DLinear \cite{dlinear} and FEDformer \cite{fedformer} in comparison with the cross-domain transfer results of DropPatch and PatchTST. We denote that \emph{PatchTST} refer to the self-supervised version PatchTST. 
We conduct experiments in both in-domain and cross-domain settings. For the in-domain setting, we pre-train and fine-tune the model using the same dataset. In the cross-domain setting, we pre-train the model on one dataset and then fine-tune it on other target datasets to evaluate its adaptability and generality across diverse scenarios.
Unless otherwise stated, the input sequence length of DropPatch is set to 512, and the patch length is fixed at 12 following the self-supervised PatchTST \cite{PatchTST}. This configuration results in a total of 42 patches.
% More implementation details are provided in Appendix \ref{sec:implementation}.

\subsubsection{Main Results} Our proposed DropPatch exhibits significant improvement over other established strong baselines in various time-series forecasting scenarios, while enjoying the computational efficienty and reduced memory usage.

\subsection{In-Domain Forecasting}
We conduct time-series forecasting experiments under an in-domain setting, where models are pre-trained and fine-tuned on the same datasets. The results are summarized in Table \ref{tab:indomain}.

\begin{table*}[h]
\centering
\caption{In-domain time-series forecasting results, averaged from all forecasting steps $T \in \{96, 192, 336, 720\}$.}
\label{tab:indomain}
\scriptsize
\begin{tabular}{c|c|c|c|c|c|c|c|c}
\toprule
Models & \textbf{DropPatch} & PatchTST & SimMTM & Ti-MAE & TST & LaST & CoST & TS2Vec \\
\midrule
Metrics & MSE ~ MAE & MSE ~ MAE & MSE ~ MAE & MSE ~ MAE & MSE ~ MAE & MSE ~ MAE & MSE ~ MAE & MSE ~ MAE \\
\midrule
ETTm1 
& \textbf{0.336 ~ 0.378} & 0.341 ~ 0.379 & 0.340 ~ 0.379 & 0.682 ~ 0.532 
& 0.494 ~ 0.471 & 0.383 ~ 0.399 & 0.477 ~ 0.486 & 0.664 ~ 0.689 \\
\midrule
ETTm2 
& \textbf{0.254 ~ 0.315} & 0.258 ~ 0.318 & 0.260 ~ 0.318 & 0.392 ~ 0.417  
& 0.425 ~ 0.371 & 0.389 ~ 0.394 & 0.825 ~ 0.651 & 0.359 ~ 0.420 \\
\midrule
ETTh1 
& \textbf{0.400} ~ 0.429 & 0.430 ~ 0.445 & 0.404 ~ \textbf{0.428} & 0.721 ~ 0.591 
& 0.624 ~ 0.562 & 0.571 ~ 0.532 & 0.710 ~ 0.627 & 0.643 ~ 0.728 \\
\midrule
ETTh2 
& \textbf{0.347 ~ 0.390} & 0.355 ~ 0.394 & 0.348 ~ 0.391 & 0.482 ~ 0.488 
& 0.429 ~ 0.458 & 0.499 ~ 0.497 & 1.664 ~ 0.999 & 0.801 ~ 0.856   \\
\midrule
Weather 
& \textbf{0.220 ~ 0.259} & 0.225 ~ 0.261 & 0.235 ~ 0.280 & 0.324 ~ 0.343 
& 0.419 ~ 0.448 & 0.237 ~ 0.268 & 1.111 ~ 0.801 & 0.658 ~ 0.751 \\
\midrule
ECL 
& \textbf{0.157 ~ 0.249} & \textbf{0.157} ~ 0.252 & 0.162 ~ 0.356 & 0.561 ~ 0.554 
& 0.310 ~ 0.353 & 0.186 ~ 0.274 & 0.228 ~ 0.335 & 0.354 ~ 0.427\\
\midrule
Traffic 
& \textbf{0.378 ~ 0.257} & 0.382 ~ 0.259 & 0.392 ~ 0.264 & 0.916 ~ 0.423 
& 0.611 ~ 0.503 & 0.713 ~ 0.397 & 0.760 ~ 0.428 & 0.501 ~ 0.375\\
\bottomrule
\end{tabular}
\end{table*}

In-domain experiments show that our DropPatch strategy surpasses existing methods in 13 out of 14 metrics across 7 datasets. Each metric demonstrates significant superiority in comparison with other baselines. PatchTST is noted as a strong baseline. Nevertheless, by simply applying the DropPatch strategy, performance is further improved in both MSE and MAE, with only half the time consumption and memory usage in pre-training stage.

The forecasting performance of PatchTST, SimMTM, and DropPatch is significantly superior to other baselines. The commonality among these three methods is the use of channel-independent masked time-series modeling. 
% For time-series forecasting tasks, masking and reconstruction is a more reasonable pre-training task compared to contrastive learning, aligning with the conclusions of Study \cite{SimMTM}.

Compared to PatchTST, the DropPatch strategy offers further improvements in this task. This is primarily because the masked time-series modeling task can be done with a little understanding of underlying patterns in the time-series, which can lead to superficial learning and over-fitting. 
Random dropping introduces a significant amount of randomness to each sample, thus acting as a data augmentation method that helps mitigate the over-fitting issue. 
In the meanwhile, the challenging pre-training task requires a comprehensive understanding of underlying patterns and thus encourages the learning of useful representation.
% Thoroughly validation and discussion are presented in Appendix \ref{sec:trainingcurve}.

\subsection{Cross-Domain Forecasting}
In this section, we explore multiple scenarios in cross-domain transfer learning. We perform fine-tuning on target datasets using all available training samples. Specifically, we conduct experiments with 1) ECL as the fixed source dataset, following the setup in \cite{PatchTST}, and 2) ETTm1 as the fixed target dataset. The results are summarized in Table \ref{tab:fullft1} \ref{tab:fullft2}. Notably, when the source dataset has a mismatch in the number of channels compared to the target dataset, some baseline models are unable to perform the transfer. Although SimMTM is capable of transferring under conditions of channel mismatch, we encountered an out-of-memory (OOM) issue when pre-training SimMTM on the ECL dataset, even with a batch size of 1. 
Therefore, we also include supervised models for comparison when using ECL as the source dataset.

From the comparison, we observe that DropPatch significantly surpasses the other baselines. Notably, while PatchTST falls behind some supervised methods, DropPatch consistently outperforms these supervised methods. The improved performance stems from the prevention of severe over-fitting in the source dataset, ensuring the model's robustness and generalization capability when applied to unseen target datasets. In contrast, over-fitting can hinder PatchTST's ability to generalize effectively to new patterns.
% This advantage mainly stems from: the DropPatch strategy disrupts the time-series patterns that are unique to a specific dataset, making the model focus more on generalized features of the time-series. To validate this hypothesis, we conducted analytical experiments in Section \ref{sec:attndis} and Section \ref{sec:attndistributsion}, which showed that the model tends to focus more on global and local patterns and that the attention coefficients are more distinctive.

\begin{table*}[ht]
\caption{Cross-domain time-series forecasting results. ECL$\rightarrow$ETTm1 denotes the models are pre-trained on ECL and then are fine-tuned on ETTm1. iTransformer, DLinear, and FEDformer are trained directly on the target dataset using supervised learning. Results are averaged from all forecasting steps $T \in \{96, 192, 336, 720\}$.}
\label{tab:fullft1}
\centering
\scriptsize
\begin{tabular}{c|c|c|c|c|c}
\toprule
Models & \textbf{DropPatch} & PatchTST & iTransformer & DLinear & FEDformer \\
\midrule
Metrics & MSE ~ MAE & MSE ~ MAE & MSE ~ MAE & MSE ~ MAE & MSE ~ MAE \\
\midrule
ECL$\rightarrow$ETTm1 & 0.349 ~ 0.383 & \textbf{0.346} ~ 0.383 & 0.371 ~ 0.400 & 0.357 ~ \textbf{0.379} & 0.382 ~ 0.422  \\
\midrule
ECL$\rightarrow$ETTm2 & 0.258 ~ 0.321 & \textbf{0.257 ~ 0.318} & 0.272 ~ 0.333 & 0.267 ~ 0.332 & 0.292 ~ 0.343  \\
\midrule
ECL$\rightarrow$ETTh1 & \textbf{0.395 ~ 0.426} & 0.434 ~ 0.448 & 0.451 ~ 0.462 & 0.423 ~ 0.437 & 0.428 ~ 0.454  \\
\midrule
ECL$\rightarrow$ETTh2 & \textbf{0.350 ~ 0.392} & 0.354 ~ 0.395 & 0.387 ~ 0.418 & 0.431 ~ 0.447 & 0.388 ~ 0.434  \\
\midrule
ECL$\rightarrow$Weather & \textbf{0.222 ~ 0.260} & 0.226 ~ 0.264 & 0.246 ~ 0.279 & 0.246 ~ 0.300 & 0.310 ~ 0.357  \\
\midrule
ECL$\rightarrow$Traffic & \textbf{0.379 ~ 0.257} & 0.411 ~ 0.285 & 0.380 ~ 0.271 & 0.434 ~ 0.295 & 0.604 ~ 0.372  \\
\bottomrule
\end{tabular}
\end{table*}

\begin{table*}[ht]
\caption{Cross-domain time-series forecasting results. ETTh1$\rightarrow$ETTm1 denotes the models are pre-trained on ETTh1 and then are fine-tuned on ETTm1. Results are averaged from all forecasting steps $T \in \{96, 192, 336, 720\}$. Notation "$-$" means transfer learning is not feasible due to the mismatch in the number of channels. }
\label{tab:fullft2}
\centering
\scriptsize
\resizebox{\textwidth}{!}{
\begin{tabular}{c|c|c|c|c|c|c|c|c|c}
\toprule
Models & \textbf{DropPatch} & PatchTST & SimMTM & Ti-MAE & TST & LaST & TF-C & CoST & TS2Vec\\
\midrule
Metrics & MSE\hspace{4pt}MAE & MSE\hspace{4pt}MAE & MSE\hspace{4pt}MAE & MSE\hspace{4pt}MAE & MSE\hspace{4pt}MAE & MSE\hspace{4pt}MAE & MSE\hspace{4pt}MAE & MSE\hspace{4pt}MAE & MSE\hspace{4pt}MAE\\
\midrule
ETTh1$\rightarrow$ETTm1 & 0.352\hspace{4pt}0.386 & 0.352\hspace{4pt}0.386 & \textbf{0.346}\hspace{4pt}\textbf{0.384} & 0.666\hspace{4pt}0.529 & 0.482\hspace{4pt}0.444 & 0.353\hspace{4pt}0.390 & 0.746\hspace{4pt}0.562 & 0.359\hspace{4pt}0.407 & 0.697\hspace{4pt}0.616\\
\midrule
ETTh2$\rightarrow$ETTm1 & \textbf{0.361}\hspace{4pt}0.390 & 0.364\hspace{4pt}0.391 & 0.365\hspace{4pt}\textbf{0.384} & 0.688\hspace{4pt}0.535 & 0.472\hspace{4pt}0.448 & 0.475\hspace{4pt}0.489 & 0.750\hspace{4pt}0.654 & 0.377\hspace{4pt}0.413 & 0.606\hspace{4pt}0.556\\
\midrule
ETTm2$\rightarrow$ETTm1 & \textbf{0.343}\hspace{4pt}\textbf{0.382} & 0.353\hspace{4pt}0.390 & 0.351\hspace{4pt}0.383 & 0.682\hspace{4pt}0.531 & 0.480\hspace{4pt}0.455 & 0.414\hspace{4pt}0.464 & 0.758\hspace{4pt}0.669 & 0.354\hspace{4pt}0.401 & 0.756\hspace{4pt}0.638\\
\midrule
Weather$\rightarrow$ETTm1 & \textbf{0.348}\hspace{4pt}\textbf{0.385} & 0.359\hspace{4pt}0.390 & 0.358\hspace{4pt}0.388 & - & - & - & - & - & -\\
\bottomrule
\end{tabular}}
\end{table*}

\subsection{Evaluations on Synthesized Dataset}
In the cross-domain experiments mentioned above, the models are initially pre-trained on a single source dataset and then fine-tuned on a target dataset. For the purpose of developing time-series foundation models \cite{moment,moirai,timer}, the source dataset could be a mixed dataset. 
In the mixed dataset, time-series samples are from different domains, exhibiting varying frequencies, and containing diverse semantic information. This setup aims to enhance the model's robustness and ability to generalize across different scenarios, while also posing a challenge for models to handle diverse data. 

We compile two synthesized datasets to facilitate multi-dataset pre-training for evaluation.
This section primarily focuses on exploring the potential of applying DropPatch to time-series foundation models, without the concern with pushing state-of-the-art results. 
% And consequently, our experiments are exclusively focused on these two methods.

Specifically, we merge 10 datasets to compile a synthesized time-series dataset, named STS66M, which has a total file size of over 66 MB and consists of more than 3.76 million data points.
The models are pre-trained on STS66M and subsequently fine-tuned on other target datasets. The averaged results are in Table \ref{tab:merge_ft}. DropPatch significantly outperforms PatchTST, demonstrating its superior adaptability to diverse pre-training data and its ability to learn more robust and general representations for downstream tasks.

An important application of pre-trained models is to provide priori knowledge for downstream datasets, particularly in scenarios with limited fine-tuning data availability, commonly referred to as few-shot learning. This capability is crucial for the fast adaptation of deep models, which has been demonstrated remarkable performance in NLP \cite{gpt3,gpt4}. 
To further explore this, we expand the size of our synthesized time-series dataset by including ECL and PEMS07. The expanded dataset has a file size over 162MB, named STS162M, consisting of 32.5 million data points. 
We then conduct few-shot learning experiments using models pre-trained on STS162M. The results are presented in Table \ref{tab:merge_fewshot}. For each unseen target dataset, we employ only the headmost 100, 300, and 500 training samples to evaluate DropPatch and PatchTST. DropPatch can generalize well and achieve improved performance.

\begin{table*}[h]
\caption{Cross-domain fine-tuning results. Models are pre-trained on STS66M, then fine-tuned on other unseen datasets. Forecasting steps $T \in \{96, 192, 336, 720\}$.}
\label{tab:merge_ft}
\centering
\scriptsize
\begin{tabular}{cc|c|c|c|c|c|c|c}
\toprule
& Datasets & Weather & ETTh1 & ETTh2 & ETTm1 & ETTm2 & ECL & Traffic\\
\midrule
Models& $S$& MSE ~ MAE& MSE ~ MAE& MSE ~ MAE& MSE ~ MAE& MSE ~ MAE& MSE ~ MAE & MSE ~ MAE\\
\midrule
\multirow{5}{*}{\textbf{DropPatch}}      
& 96   & \textbf{0.142 ~ 0.190} & \textbf{0.374 ~ 0.409} & \textbf{0.288 ~ 0.346} & \textbf{0.289 ~ 0.345} & 0.171 ~ 0.261 & \textbf{0.129 ~ 0.221} & \textbf{0.361 ~ 0.255}  \\
& 192  & \textbf{0.186 ~ 0.234} & \textbf{0.401 ~ 0.427} & \textbf{0.352 ~ 0.385} & \textbf{0.334 ~ 0.373} & \textbf{0.229 ~ 0.301} & \textbf{0.148 ~ 0.239} &\textbf{ 0.378 ~ 0.262}  \\
& 336  & \textbf{0.238 ~ 0.274} & \textbf{0.406} ~ 0.437 & \textbf{0.360 ~ 0.401} &\textbf{ 0.361 ~ 0.394} & 0.282 ~ 0.337 &\textbf{ 0.165 ~ 0.258} & \textbf{0.389 ~ 0.268}  \\
& 720  &\textbf{ 0.312 ~ 0.330} & 0.446 ~ 0.469 & \textbf{0.384 ~ 0.426} & \textbf{0.408 ~ 0.426} & \textbf{0.365} ~ 0.389 & \textbf{0.201 ~ 0.290} & \textbf{0.427 ~ 0.289}  \\
& AVG  & \textbf{0.220 ~ 0.257} & \textbf{0.407} ~ 0.436 & \textbf{0.346 ~ 0.390} & \textbf{0.348 ~ 0.385} & \textbf{0.262 ~ 0.322} & \textbf{0.161 ~ 0.252} & \textbf{0.389 ~ 0.269} \\
\midrule
\multirow{5}{*}{PatchTST} 
& 96   & 0.144 ~ 0.193 & 0.381 ~ 0.412 & 0.303 ~ 0.355 & 0.293 ~ 0.346 & \textbf{0.170 ~ 0.262} & 0.131 ~ 0.224 & 0.372 ~ 0.266  \\
& 192  & 0.191 ~ 0.240 & 0.407 ~ 0.430 & 0.367 ~ 0.390 & 0.336 ~ 0.375 & 0.235 ~ 0.309 & \textbf{0.148} ~ 0.240 & 0.389 ~ 0.272  \\
& 336  & 0.244 ~ 0.281 & 0.411 ~ \textbf{0.435} & 0.366 ~ 0.403 & 0.364 ~ \textbf{0.394} & \textbf{0.280 ~ 0.334} & \textbf{0.165 ~ 0.258} & 0.396 ~ 0.273  \\
& 720  & 0.317 ~ 0.334 & \textbf{0.443 ~ 0.464} & 0.395 ~ 0.431 & 0.412 ~ 0.428 & 0.366 ~ \textbf{0.387} & 0.203 ~ 0.291 & 0.434 ~ 0.293  \\
& AVG  & 0.224 ~ 0.262 & 0.411 ~ \textbf{0.435} & 0.358 ~ 0.395 & 0.351 ~ 0.386 & 0.263 ~ 0.323 & 0.162 ~ 0.253 & 0.398 ~ 0.276  \\
\bottomrule
\end{tabular}
\end{table*}

\begin{table*}[h]
\caption{Few-shot learning results. Models are pre-trained on STS162M, then fine-tuned on other unseen datasets using limited training samples. Forecasting steps is fixed at 96.}
\label{tab:merge_fewshot}
\centering
\scriptsize
\begin{tabular}{cc|c|c|c|c|c|c}
\toprule
& Datasets & Weather & ETTh1 & ETTh2 & ETTm1 & ETTm2 & Traffic\\
\midrule
Models& \# Samples& MSE ~ MAE& MSE ~ MAE& MSE ~ MAE& MSE ~ MAE& MSE ~ MAE& MSE ~ MAE \\
\midrule
\multirow{3}{*}{\textbf{DropPatch}}
& 100  & \textbf{0.242 ~ 0.290} & \textbf{0.626 ~ 0.525}  & \textbf{0.372} ~ 0.411  & \textbf{0.502 ~ 0.465}  & \textbf{0.277 ~ 0.343}  & \textbf{0.447 ~ 0.309}  \\
& 300  & 0.223 ~ 0.273  & \textbf{0.506 ~ 0.488}  & \textbf{0.312 ~ 0.366}  & 0.531 ~ 0.478  & \textbf{0.237 ~ 0.311} & \textbf{0.399 ~ 0.275}  \\
& 500  & \textbf{0.212 ~ 0.263}  & \textbf{0.474 ~ 0.461}  & \textbf{0.317 ~ 0.366}  & 0.518 ~ 0.476  & \textbf{0.210 ~ 0.292}  & \textbf{0.395 ~ 0.275}  \\
\midrule
\multirow{3}{*}{PatchTST} 
& 100  & 0.247 ~ 0.294  & 0.666 ~ 0.552  & 0.381 ~ \textbf{0.401}  & 0.521 ~ 0.474  & 0.282 ~ 0.347  & 0.450 ~ 0.313  \\
& 300  & \textbf{0.222 ~ 0.271}  & 0.520 ~ 0.503  & 0.319 ~ 0.375  & \textbf{0.508 ~ 0.469}  & 0.257 ~ 0.327  & \textbf{0.399} ~ 0.276  \\
& 500  & 0.225 ~ 0.274  & 0.483 ~ 0.481  & 0.323 ~ 0.372  & \textbf{0.493 ~ 0.461}  & 0.214 ~ 0.298  & 0.396 ~ \textbf{0.275}  \\
\bottomrule
\end{tabular}
\end{table*}

% \textbf{Few Shot}. An important application of pre-trained models is to provide a priori knowledge for downstream datasets, particularly in scenarios with limited data availability, known as few shot learning. This capability is crucial for the fast adaptation of deep models, which has been demonstrated remarkable performance in NLP\cite{gpt3,gpt4}. 
% Using a model pre-trained on the Electricity dataset, we conduct few-shot learning on other datasets. Consistent with conventional experimental settings, we partition these downstream datasets into training, validation, and test sets. However, during the fine-tuning stage, only 5\% and 10\% of the training data (time points) are utilized. 
% The results are shown in Table \ref{tab:fewshot}. 
% Few-shot transfer learning presents a challenging task for time-series forecasting, with results significantly degraded compared to those achieved under full fine-tuning setting. Nonetheless, the forecasting performance of DropPatch remains superior to that of other baseline models.

\begin{table*}[ht]
  \begin{minipage}{0.4\linewidth}
    \centering
    \caption{Results of cold start setup. The lookback length $L_{ft}$ is fixed at 96. Results are averaged from all forecasting steps. }
    \label{tab:coldstart}
    \centering
    \scriptsize
    \begin{tabular}{c|c|c}
    \toprule
    Models & \textbf{DropPatch} & PatchTST \\
    \midrule
    Metrics & MSE MAE & MSE MAE \\
    \midrule
    Exchange & \textbf{0.348 ~ 0.396} & 0.354 ~ 0.400 \\
    PEMS03 & \textbf{0.198 ~ 0.293} &0.205 ~ 0.296 \\
    PEMS04 & \textbf{0.264 ~ 0.339} &0.273 ~ 0.343 \\
    PEMS07 & \textbf{0.214 ~ 0.312} &0.219 ~ 0.323 \\
    PEMS08 & \textbf{0.225 ~ 0.300} &0.233 ~ 0.305 \\
    \bottomrule
    \end{tabular}
  \end{minipage}
  \hfill
  \begin{minipage}{0.5\linewidth}
    \centering
    \caption{Model efficiency comparison. \textit{Mem.} denotes the memory usage, measured in megabytes (MB). \textit{T.C.} denotes the time consumption per epoch in seconds.}
    \label{tab:model_efficiency}
    \centering
    \scriptsize
    \begin{tabular}{c|c|c|c}
    \toprule
    Models & \textbf{DropPatch} & PatchTST & SimMTM\\
    \midrule
    Metrics & Mem. ~ T.C. & Mem. ~ T.C. & Mem. ~ T.C.\\
    \midrule
    ETTm1 
    & \textbf{1404 ~ 32.2} & 1722 ~ 44.5 & 29090 ~ 823.3 \\
    Weather 
    & \textbf{2094 ~ 42.1}& 3914 ~ 75.1 & OOM \\
    ECL 
    & \textbf{4256 ~ 306.7} & 11050 ~ 528.5 & OOM \\
    \bottomrule
    \end{tabular}
  \end{minipage}
\end{table*}

% \begin{table}[h]
% \caption{Cold start setting. Models are pre-trained on the ECL dataset and fine-tuned on Exchange and 4 PEMSs. The lookback length $L_{ft}$ is fixed to be 96. Results are averaged from all forecasting steps. Full results are in Appendix \ref{sec:full_results_cold}.}
% \label{tab:coldstart}
% \centering
% \scriptsize
% \begin{tabular}{c|c|c}
% \toprule
% Models & \textbf{DropPatch} & PatchTST \\
% \midrule
% Metrics & MSE MAE & MSE MAE \\
% \midrule
% Exchange & \textbf{0.348 ~ 0.396} & 0.354 ~ 0.400 \\
% PEMS03 & \textbf{0.198 ~ 0.293} &0.205 ~ 0.296 \\
% PEMS04 & \textbf{0.264 ~ 0.339} &0.273 ~ 0.343 \\
% PEMS07 & \textbf{0.214 ~ 0.312} &0.219 ~ 0.323 \\
% PEMS08 & \textbf{0.225 ~ 0.300} &0.233 ~ 0.305 \\
% \bottomrule
% \end{tabular}
% \end{table}
% \subsection{Model Analysis}

\subsection{Cold Start} This task aims to forecast in target datasets where lookback $L_{ft}$ is relatively short, providing limited historical information for fine-tuning. The experimental setup was first introduced in time-series forecasting by \cite{attnshare}. In our experiments, the lookback length is fixed at $L_{ft}=96$, which is shorter than the lookback length $L_{pt}=512$ on the pre-training stage. We perform experiments on Exchange and four PEMS(PEMS03, PEMS04, PEMS07, PEMS08) as the target datasets. The source dataset is fixed as ECL. Forecasting steps $T \in \{96, 192, 336, 720\}$ for Exchange and $T \in \{12, 24, 48, 96\}$ for the PEMS datasets. We denote that under cold start scenario, the pre-trained models are expected to leverage the limited historical information for future forecasting. In Table \ref{tab:coldstart}, we present the averaged results across the target datasets. Our method consistently outperforms the baseline methods.

\subsection{Model Efficiency}
\label{sec:model_efficiency}
We compared the training speed and memory usage during the pre-training stage, results are presented in Table \ref{tab:model_efficiency}.
All experiments are conducted on a single NVIDIA Tesla V100-SXM2-32GB GPU. In comparison with the other two leading masked time-series modeling methods, DropPatch significantly reduces the memory usage and training time consumption by a large margin. This computational efficiency makes it feasible to scale up and potentially improve model performance by exposing the model to a larger dataset.

\section{Discussion}\label{sec:diss_mt}
Since its inception, the self-supervised PatchTST, which employs a patch-level masking pre-training paradigm, has consistently achieved state-of-the-art performance. Our proposed method DropPatch improves upon this by dropping a certain proportion of patches prior to applying the patch-level masking strategy, resulting in superior performance in both in-domain and cross-domain scenarios. This raises several questions: How does DropPatch strategy differ from PatchTST, and what drives its enhanced performance?

In the main text, we will provide a brief description and present the findings for each empirical study. Similar results are observed across various datasets; results on ETTm1 is displayed here as a representative example. Unless otherwise specified, the experiments are conducted in an in-domain scenario using the ETTm1 dataset.

\begin{figure*}[ht]
  \centering
  \includegraphics[width=\textwidth]{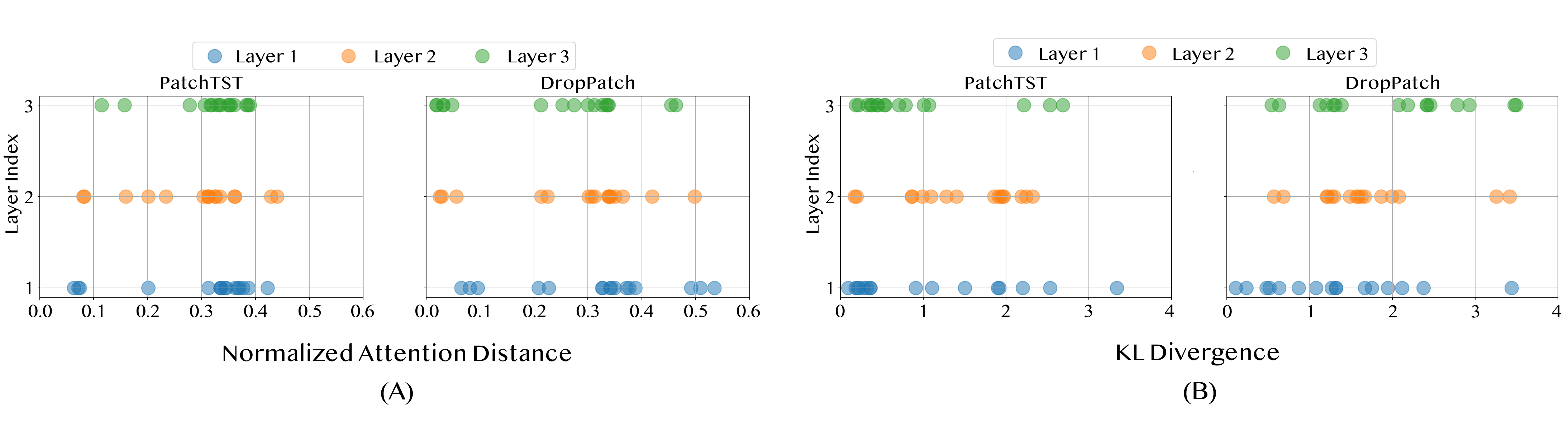}
  \caption{Analysis of (A) normalized distance, and (B) KL divergence between attention distributions and uniform distribution for each head across all layers. Each dot represents an individual attention head, while different colors indicate different layers.}
  \label{fig:attn}
\end{figure*}

% \begin{figure*}[h]
%   \centering
%   \includegraphics[width=\textwidth]{./discuss_part.pdf}
%   \caption{(A) Attention distribution differences across multi-head attention at the last layer. (B) Central Kernel Alignment Analysis between representations before and after fine-tuning.}
%   \label{fig:dicuss_part}
% \end{figure*}

\begin{figure}[h]
  \centering
  \includegraphics[width=0.5\textwidth]{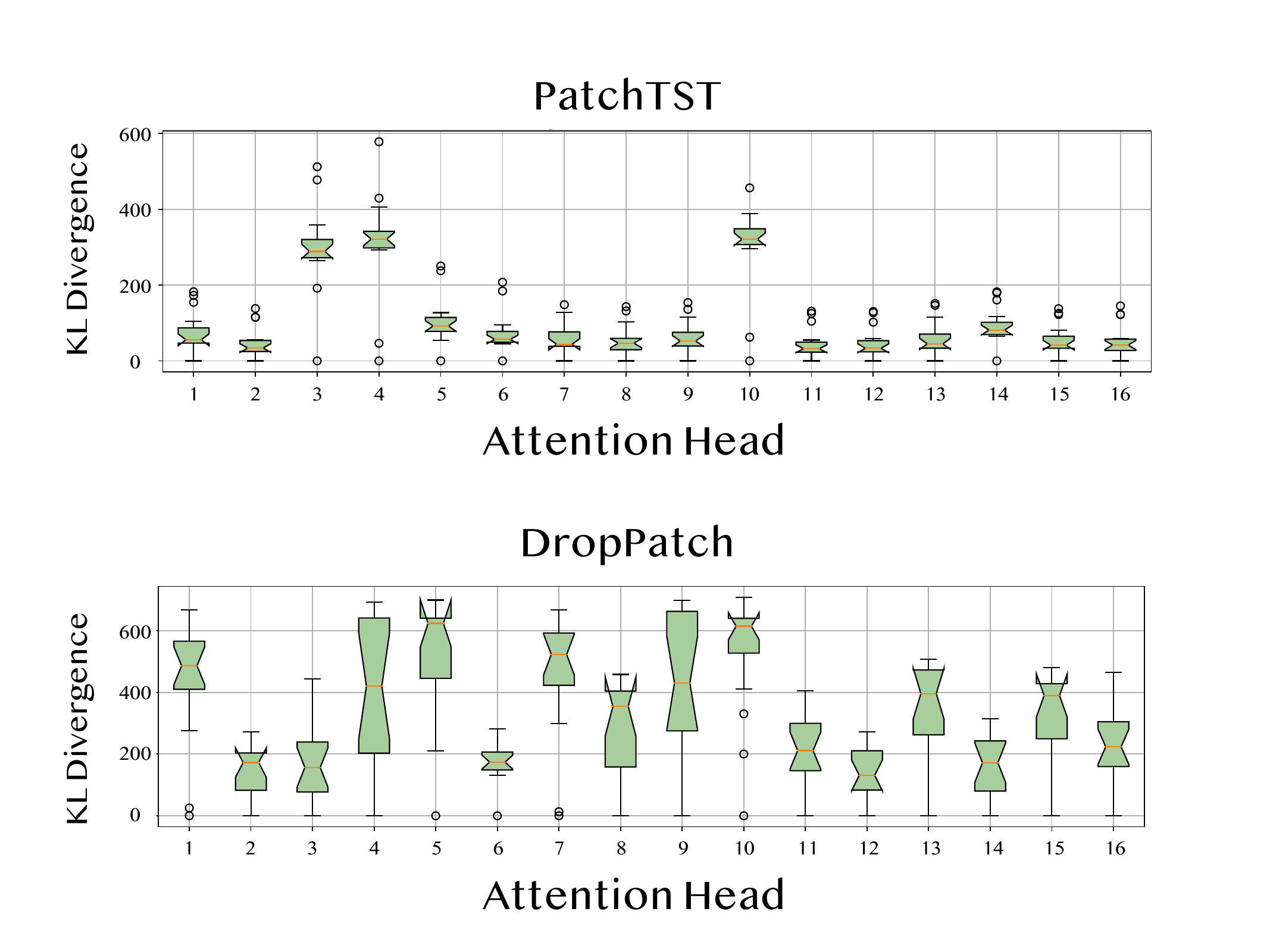}
  \caption{Attention distribution difference across attention heads in the last layer.}
  \label{fig:dicuss_part_3}
\end{figure}

\begin{figure}[h]
  \centering
  \includegraphics[width=0.5\textwidth]{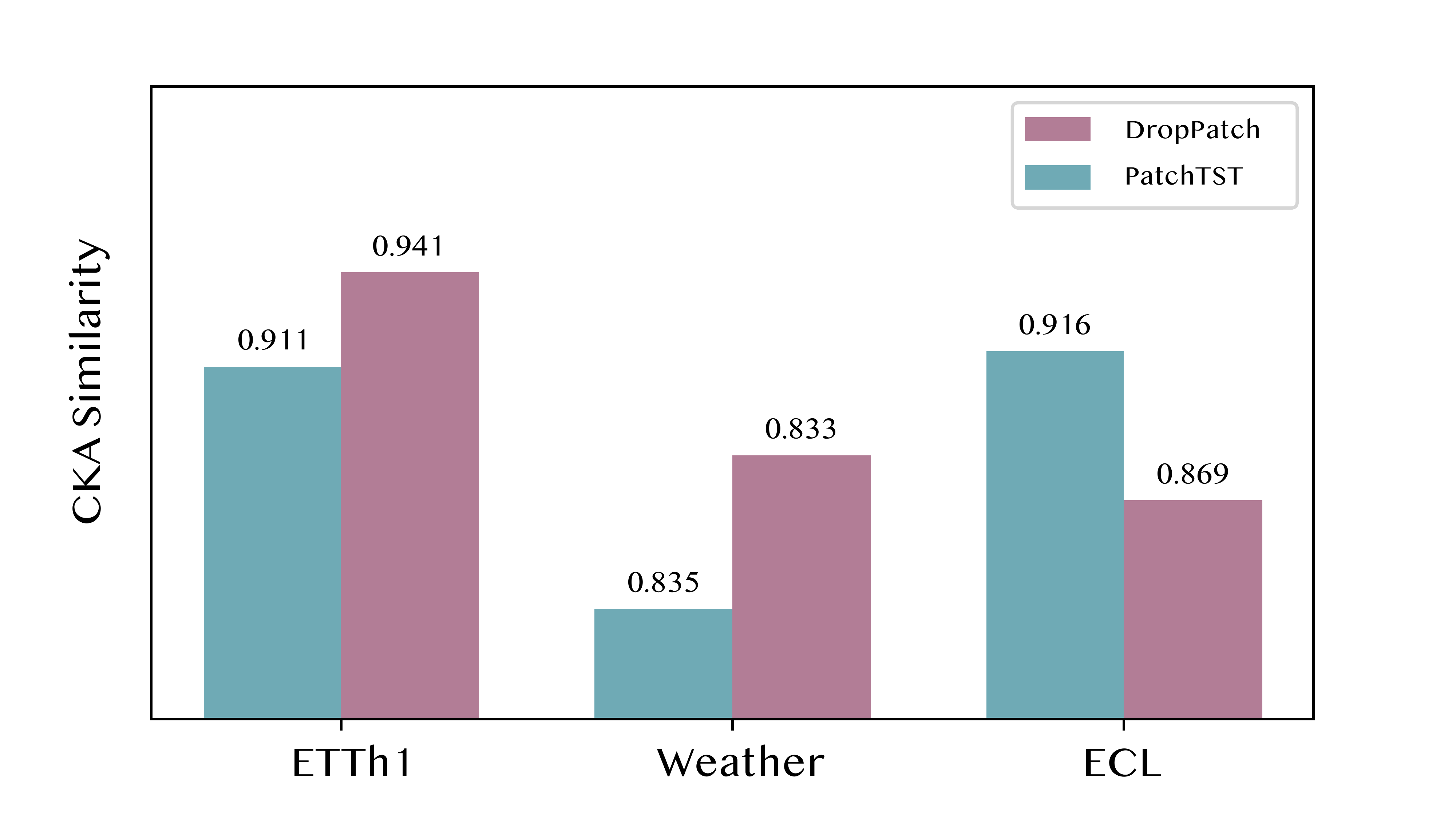}
  \caption{Models are pre-trained on the ECL dataset and subsequently fine-tuned on ECL (in-domain) and on the ETTh1 and Weather (cross-domain) datasets.}
  \label{fig:dicuss_part_4}
\end{figure}

\subsection{Normalized Attention Distance}\label{sec:attndis}

Firstly, we analyze the averaged attention distances before and after applying the DropPatch strategy. Specifically, following previous work \cite{xie2023revealing}, we define \textit{distance} as the absolute position difference between two patches, and \textit{normalized attention distance} as the product of these attention distances with the attention weights. Intuitively, a larger normalized attention distance indicate a focus on global information, while a smaller one reflect attention to local information.
The results for each head in all layers are shown in Figure \ref{fig:attn} (A).

\subsubsection{Finding 1}: By comparing normalized attention distances, we found that the DropPatch strategy enables each attention head in the model to focus on information at varying scales.
Specifically, this strategy enhancing the model's ability to capture both short-term and long-term dependencies, empowering the model with a more comprehensive understanding of the time-series.
% enhancing the distinctiveness and robustness of the representations learned by the model. 

\subsection{Attention Coefficients Distribution} \label{sec:attndistributsion}
We then analyze the distributions of attention coefficients across different heads and layers. Uniform attention coefficients lead to a loss of distinctiveness, effectively diminishing the model's ability to capture unique patterns. In contrast, distributions with sharper focus and higher distinctiveness are regarded as more effective \cite{Informer,RelTransformer,long_tailed,long_tailed2}. 
In our empirical study, we quantify the distinctiveness of these distributions by computing the Kullback-Leibler (KL) divergence between the uniform distribution and the attention distributions. A larger KL divergence indicates a greater deviation from the uniform distribution, reflecting sharper and more distinctive attention patterns. The results are shown in Figure \ref{fig:attn} (B).

\subsubsection{Finding 2}: The results indicate that applying the DropPatch strategy sharpens the focus of attention heads, facilitating the identification of more valuable information and underlying patterns.
% This more focus attention can significantly benefits the model's representation capabilities.

\subsection{Attention Coefficients Difference} 
The previous two subsections reveal that attention heads in DropPatch exhibit greater diversity in behavior. In this subsection, we further investigate whether different attention heads capture diverse information. Specifically, we conduct an analysis of the attention distribution across different heads by calculating the KL divergence between attention heads in the same layer. This comparison highlights the distributional differences among attention heads. A higher KL divergence indicates greater differences, suggesting that each head has learned distinct information, thereby reducing redundancy in the information captured by different heads. As shown in Figure \ref{fig:dicuss_part_3}, attention heads in DropPatch exhibit higher KL divergence compared to those in PatchTST.

\subsubsection{Finding 3}: The analysis of attention distributions demonstrates that the DropPatch strategy enables attention heads to capture distinct information, thereby reducing redundancy and enhancing the model's representation capabilities.

\subsection{Central Kernel Alignment Analysis} 
We use CKA (Central Kernel Alignment) values \cite{cka} to compare the similarity of representations in a pre-trained model before and after downstream fine-tuning. Specifically, we calculate CKA similarity using the last layer representations between the pre-trained model and the fine-tuned model.
% Specifically, for in-domain tasks, we aim for a high CKA similarity, which indicates that the model has sufficiently learned the underlying characteristics of the current dataset during pre-training. Conversely, a high CKA similarity in a cross-domain scenario suggests a hindrance to the model's ability to effectively generalize to new patterns. 

\subsubsection{Finding 4}: From the results as shown in Figure \ref{fig:dicuss_part_4}, we found that DropPatch strategy significantly enhances the representation ability. For in-domain tasks, DropPatch achieves high CKA similarity, indicating that the model better learns the underlying patterns of the dataset. For cross-domain tasks, DropPatch exhibits reduced CKA similarity, which we attribute to the model's improved ability to handle domain shifts and adapt to unseen distributions after applying the DropPatch strategy.

% These improvements are attributed to the random perturbations and data augmentation introduced by DropPatch, which prevent the model from superficial learning while capturing richer data, thereby enhancing the generalization capability and robustness of the representations.

% More analysis on the methods can be found in Appendix \ref{sec:attn_heatmap} \ref{sec:trainingcurve}.

\section{Conclusion}
In this paper, we propose DropPatch, an enhancement to masked time-seires modeling achieved by introducing the random dopping of sub-series patches. This approach yields significant improvements in pre-training efficiency and various downstream tasks. Extensive experiments validate the effectiveness, highlighting its ability to improve the attention mechanism by enabling a sharper focus on multi-scale and diverse information. Furthermore, out theoretical analysis reveals that this technique slows the degeneration of Transformer representations toward a rank-1 linear subspace, underlying its beneficial impact on model performance.

\bibliography{aaai25}

\begin{thebibliography}{48}
\providecommand{\natexlab}[1]{#1}

\bibitem[{Achiam et~al.(2023)Achiam, Adler, Agarwal, Ahmad, Akkaya, Aleman, Almeida, Altenschmidt, Altman, Anadkat et~al.}]{gpt4}
Achiam, J.; Adler, S.; Agarwal, S.; Ahmad, L.; Akkaya, I.; Aleman, F.~L.; Almeida, D.; Altenschmidt, J.; Altman, S.; Anadkat, S.; et~al. 2023.
\newblock Gpt-4 technical report.
\newblock \emph{arXiv preprint arXiv:2303.08774}.

\bibitem[{Amos, Berant, and Gupta(2023)}]{nevertrainfrom}
Amos, I.; Berant, J.; and Gupta, A. 2023.
\newblock Never Train from Scratch: Fair Comparison of Long-Sequence Models Requires Data-Driven Priors.
\newblock \emph{arXiv preprint arXiv:2310.02980}.

\bibitem[{Baevski et~al.(2022)Baevski, Hsu, Xu, Babu, Gu, and Auli}]{Data2VEC}
Baevski, A.; Hsu, W.-N.; Xu, Q.; Babu, A.; Gu, J.; and Auli, M. 2022.
\newblock Data2vec: A general framework for self-supervised learning in speech, vision and language.
\newblock In \emph{International Conference on Machine Learning}, 1298--1312. PMLR.

\bibitem[{Bao et~al.(2021)Bao, Dong, Piao, and Wei}]{BeiT}
Bao, H.; Dong, L.; Piao, S.; and Wei, F. 2021.
\newblock Beit: Bert pre-training of image transformers.
\newblock \emph{arXiv preprint arXiv:2106.08254}.

\bibitem[{Brown et~al.(2020)Brown, Mann, Ryder, Subbiah, Kaplan, Dhariwal, Neelakantan, Shyam, Sastry, Askell et~al.}]{gpt3}
Brown, T.; Mann, B.; Ryder, N.; Subbiah, M.; Kaplan, J.~D.; Dhariwal, P.; Neelakantan, A.; Shyam, P.; Sastry, G.; Askell, A.; et~al. 2020.
\newblock Language models are few-shot learners.
\newblock \emph{Advances in neural information processing systems}, 33: 1877--1901.

\bibitem[{Candanedo(2017)}]{appliance_energy}
Candanedo, L. 2017.
\newblock {Appliances Energy Prediction}.
\newblock UCI Machine Learning Repository.
\newblock {DOI}: https://doi.org/10.24432/C5VC8G.

\bibitem[{Candanedo, Feldheim, and Deramaix(2017)}]{applyenergy}
Candanedo, L.~M.; Feldheim, V.; and Deramaix, D. 2017.
\newblock Data driven prediction models of energy use of appliances in a low-energy house.
\newblock \emph{Energy and buildings}, 140: 81--97.

\bibitem[{Chen et~al.(2022)Chen, Agarwal, Abdelkarim, Zhu, and Elhoseiny}]{RelTransformer}
Chen, J.; Agarwal, A.; Abdelkarim, S.; Zhu, D.; and Elhoseiny, M. 2022.
\newblock Reltransformer: A transformer-based long-tail visual relationship recognition.
\newblock In \emph{Proceedings of the IEEE/CVF Conference on Computer Vision and Pattern Recognition}, 19507--19517.

\bibitem[{Cheng et~al.(2023)Cheng, Liu, Liu, Zhang, Zhang, and Chen}]{timemae}
Cheng, M.; Liu, Q.; Liu, Z.; Zhang, H.; Zhang, R.; and Chen, E. 2023.
\newblock Timemae: Self-supervised representations of time series with decoupled masked autoencoders.
\newblock \emph{arXiv preprint arXiv:2303.00320}.

\bibitem[{Choromanski et~al.(2020)Choromanski, Likhosherstov, Dohan, Song, Gane, Sarlos, Hawkins, Davis, Mohiuddin, Kaiser et~al.}]{long_tailed2}
Choromanski, K.; Likhosherstov, V.; Dohan, D.; Song, X.; Gane, A.; Sarlos, T.; Hawkins, P.; Davis, J.; Mohiuddin, A.; Kaiser, L.; et~al. 2020.
\newblock Rethinking attention with performers.
\newblock \emph{arXiv preprint arXiv:2009.14794}.

\bibitem[{Devlin et~al.(2018)Devlin, Chang, Lee, and Toutanova}]{BERT}
Devlin, J.; Chang, M.-W.; Lee, K.; and Toutanova, K. 2018.
\newblock Bert: Pre-training of deep bidirectional transformers for language understanding.
\newblock \emph{arXiv preprint arXiv:1810.04805}.

\bibitem[{Dong et~al.(2024)Dong, Wu, Zhang, Zhang, Wang, and Long}]{SimMTM}
Dong, J.; Wu, H.; Zhang, H.; Zhang, L.; Wang, J.; and Long, M. 2024.
\newblock Simmtm: A simple pre-training framework for masked time-series modeling.
\newblock \emph{Advances in Neural Information Processing Systems}, 36.

\bibitem[{Dong, Cordonnier, and Loukas(2021)}]{AttentionColl2}
Dong, Y.; Cordonnier, J.-B.; and Loukas, A. 2021.
\newblock Attention is not all you need: Pure attention loses rank doubly exponentially with depth.
\newblock In \emph{International Conference on Machine Learning}, 2793--2803. PMLR.

\bibitem[{Godahewa et~al.(2021)Godahewa, Bergmeir, Webb, Hyndman, and Montero-Manso}]{monash}
Godahewa, R.; Bergmeir, C.; Webb, G.~I.; Hyndman, R.~J.; and Montero-Manso, P. 2021.
\newblock Monash time series forecasting archive.
\newblock \emph{arXiv preprint arXiv:2105.06643}.

\bibitem[{Goswami et~al.(2024)Goswami, Szafer, Choudhry, Cai, Li, and Dubrawski}]{moment}
Goswami, M.; Szafer, K.; Choudhry, A.; Cai, Y.; Li, S.; and Dubrawski, A. 2024.
\newblock MOMENT: A Family of Open Time-series Foundation Models.
\newblock \emph{arXiv preprint arXiv:2402.03885}.

\bibitem[{He et~al.(2022)He, Chen, Xie, Li, Doll{\'a}r, and Girshick}]{MaskedAE}
He, K.; Chen, X.; Xie, S.; Li, Y.; Doll{\'a}r, P.; and Girshick, R. 2022.
\newblock Masked autoencoders are scalable vision learners.
\newblock In \emph{Proceedings of the IEEE/CVF conference on computer vision and pattern recognition}, 16000--16009.

\bibitem[{He et~al.(2020)He, Fan, Wu, Xie, and Girshick}]{moco}
He, K.; Fan, H.; Wu, Y.; Xie, S.; and Girshick, R. 2020.
\newblock Momentum contrast for unsupervised visual representation learning.
\newblock In \emph{Proceedings of the IEEE/CVF conference on computer vision and pattern recognition}, 9729--9738.

\bibitem[{Hendrycks and Gimpel(2016)}]{gelu}
Hendrycks, D.; and Gimpel, K. 2016.
\newblock Gaussian error linear units (gelus).
\newblock \emph{arXiv preprint arXiv:1606.08415}.

\bibitem[{Hogue(2019)}]{metrovolume}
Hogue, J. 2019.
\newblock {Metro Interstate Traffic Volume}.
\newblock UCI Machine Learning Repository.
\newblock {DOI}: https://doi.org/10.24432/C5X60B.

\bibitem[{Jin et~al.(2022)Jin, Park, Maddix, Wang, and Wang}]{attnshare}
Jin, X.; Park, Y.; Maddix, D.; Wang, H.; and Wang, Y. 2022.
\newblock Domain adaptation for time series forecasting via attention sharing.
\newblock In \emph{International Conference on Machine Learning}, 10280--10297. PMLR.

\bibitem[{Kornblith et~al.(2019)Kornblith, Norouzi, Lee, and Hinton}]{cka}
Kornblith, S.; Norouzi, M.; Lee, H.; and Hinton, G. 2019.
\newblock Similarity of neural network representations revisited.
\newblock In \emph{International conference on machine learning}, 3519--3529. PMLR.

\bibitem[{Liang et~al.(2024)Liang, Wen, Nie, Jiang, Jin, Song, Pan, and Wen}]{Survey1}
Liang, Y.; Wen, H.; Nie, Y.; Jiang, Y.; Jin, M.; Song, D.; Pan, S.; and Wen, Q. 2024.
\newblock Foundation Models for Time Series Analysis: A Tutorial and Survey.
\newblock \emph{arXiv preprint arXiv:2403.14735}.

\bibitem[{Liu et~al.(2022)Liu, Zeng, Chen, Xu, Lai, Ma, and Xu}]{scinet}
Liu, M.; Zeng, A.; Chen, M.; Xu, Z.; Lai, Q.; Ma, L.; and Xu, Q. 2022.
\newblock Scinet: Time series modeling and forecasting with sample convolution and interaction.
\newblock \emph{Advances in Neural Information Processing Systems}, 35: 5816--5828.

\bibitem[{Liu et~al.(2023)Liu, Hu, Zhang, Wu, Wang, Ma, and Long}]{itransformer}
Liu, Y.; Hu, T.; Zhang, H.; Wu, H.; Wang, S.; Ma, L.; and Long, M. 2023.
\newblock itransformer: Inverted transformers are effective for time series forecasting.
\newblock \emph{arXiv preprint arXiv:2310.06625}.

\bibitem[{Liu et~al.(2019)Liu, Ott, Goyal, Du, Joshi, Chen, Levy, Lewis, Zettlemoyer, and Stoyanov}]{ROBERTa}
Liu, Y.; Ott, M.; Goyal, N.; Du, J.; Joshi, M.; Chen, D.; Levy, O.; Lewis, M.; Zettlemoyer, L.; and Stoyanov, V. 2019.
\newblock Roberta: A robustly optimized bert pretraining approach.
\newblock \emph{arXiv preprint arXiv:1907.11692}.

\bibitem[{Liu et~al.(2024)Liu, Zhang, Li, Huang, Wang, and Long}]{timer}
Liu, Y.; Zhang, H.; Li, C.; Huang, X.; Wang, J.; and Long, M. 2024.
\newblock Timer: Transformers for Time Series Analysis at Scale.
\newblock \emph{arXiv preprint arXiv:2402.02368}.

\bibitem[{Nie et~al.(2022)Nie, Nguyen, Sinthong, and Kalagnanam}]{PatchTST}
Nie, Y.; Nguyen, N.~H.; Sinthong, P.; and Kalagnanam, J. 2022.
\newblock A time series is worth 64 words: Long-term forecasting with transformers.
\newblock \emph{arXiv preprint arXiv:2211.14730}.

\bibitem[{Noci et~al.(2022)Noci, Anagnostidis, Biggio, Orvieto, Singh, and Lucchi}]{AttentionColl1}
Noci, L.; Anagnostidis, S.; Biggio, L.; Orvieto, A.; Singh, S.~P.; and Lucchi, A. 2022.
\newblock Signal propagation in transformers: Theoretical perspectives and the role of rank collapse.
\newblock \emph{Advances in Neural Information Processing Systems}, 35: 27198--27211.

\bibitem[{Paszke et~al.(2019)Paszke, Gross, Massa, Lerer, Bradbury, Chanan, Killeen, Lin, Gimelshein, Antiga et~al.}]{pytorch}
Paszke, A.; Gross, S.; Massa, F.; Lerer, A.; Bradbury, J.; Chanan, G.; Killeen, T.; Lin, Z.; Gimelshein, N.; Antiga, L.; et~al. 2019.
\newblock Pytorch: An imperative style, high-performance deep learning library.
\newblock \emph{Advances in neural information processing systems}, 32.

\bibitem[{Radford et~al.(2019)Radford, Wu, Child, Luan, Amodei, Sutskever et~al.}]{gpt}
Radford, A.; Wu, J.; Child, R.; Luan, D.; Amodei, D.; Sutskever, I.; et~al. 2019.
\newblock Language models are unsupervised multitask learners.
\newblock \emph{OpenAI blog}, 1(8): 9.

\bibitem[{Raffel et~al.(2020)Raffel, Shazeer, Roberts, Lee, Narang, Matena, Zhou, Li, and Liu}]{raffel2020exploring}
Raffel, C.; Shazeer, N.; Roberts, A.; Lee, K.; Narang, S.; Matena, M.; Zhou, Y.; Li, W.; and Liu, P.~J. 2020.
\newblock Exploring the limits of transfer learning with a unified text-to-text transformer.
\newblock \emph{Journal of machine learning research}, 21(140): 1--67.

\bibitem[{Salam and El~Hibaoui(2023)}]{pctc}
Salam, A.; and El~Hibaoui, A. 2023.
\newblock {Power Consumption of Tetouan City}.
\newblock UCI Machine Learning Repository.
\newblock {DOI}: https://doi.org/10.24432/C5B034.

\bibitem[{Vaswani et~al.(2017)Vaswani, Shazeer, Parmar, Uszkoreit, Jones, Gomez, Kaiser, and Polosukhin}]{Transformer}
Vaswani, A.; Shazeer, N.; Parmar, N.; Uszkoreit, J.; Jones, L.; Gomez, A.~N.; Kaiser, {\L}.; and Polosukhin, I. 2017.
\newblock Attention is all you need.
\newblock \emph{Advances in neural information processing systems}, 30.

\bibitem[{Vito(2016)}]{air_quality}
Vito, S. 2016.
\newblock {Air Quality}.
\newblock UCI Machine Learning Repository.
\newblock {DOI}: https://doi.org/10.24432/C59K5F.

\bibitem[{Vyas, Katharopoulos, and Fleuret(2020)}]{long_tailed}
Vyas, A.; Katharopoulos, A.; and Fleuret, F. 2020.
\newblock Fast transformers with clustered attention.
\newblock \emph{Advances in Neural Information Processing Systems}, 33: 21665--21674.

\bibitem[{Wang et~al.(2022)Wang, Xu, Zhang, Trajcevski, Zhong, and Zhou}]{last}
Wang, Z.; Xu, X.; Zhang, W.; Trajcevski, G.; Zhong, T.; and Zhou, F. 2022.
\newblock Learning latent seasonal-trend representations for time series forecasting.
\newblock \emph{Advances in Neural Information Processing Systems}, 35: 38775--38787.

\bibitem[{Woo et~al.(2024)Woo, Liu, Kumar, Xiong, Savarese, and Sahoo}]{moirai}
Woo, G.; Liu, C.; Kumar, A.; Xiong, C.; Savarese, S.; and Sahoo, D. 2024.
\newblock Unified training of universal time series forecasting transformers.
\newblock \emph{arXiv preprint arXiv:2402.02592}.

\bibitem[{Woo et~al.(2022)Woo, Liu, Sahoo, Kumar, and Hoi}]{cost}
Woo, G.; Liu, C.; Sahoo, D.; Kumar, A.; and Hoi, S. 2022.
\newblock Cost: Contrastive learning of disentangled seasonal-trend representations for time series forecasting.
\newblock \emph{arXiv preprint arXiv:2202.01575}.

\bibitem[{Wu et~al.(2021)Wu, Xu, Wang, and Long}]{autoformer}
Wu, H.; Xu, J.; Wang, J.; and Long, M. 2021.
\newblock Autoformer: Decomposition transformers with auto-correlation for long-term series forecasting.
\newblock \emph{Advances in neural information processing systems}, 34: 22419--22430.

\bibitem[{Xie et~al.(2023)Xie, Geng, Hu, Zhang, Hu, and Cao}]{xie2023revealing}
Xie, Z.; Geng, Z.; Hu, J.; Zhang, Z.; Hu, H.; and Cao, Y. 2023.
\newblock Revealing the dark secrets of masked image modeling.
\newblock In \emph{Proceedings of the IEEE/CVF conference on computer vision and pattern recognition}, 14475--14485.

\bibitem[{Xie et~al.(2022)Xie, Zhang, Cao, Lin, Bao, Yao, Dai, and Hu}]{simmim}
Xie, Z.; Zhang, Z.; Cao, Y.; Lin, Y.; Bao, J.; Yao, Z.; Dai, Q.; and Hu, H. 2022.
\newblock Simmim: A simple framework for masked image modeling.
\newblock In \emph{Proceedings of the IEEE/CVF conference on computer vision and pattern recognition}, 9653--9663.

\bibitem[{Yue et~al.(2022)Yue, Wang, Duan, Yang, Huang, Tong, and Xu}]{ts2vec}
Yue, Z.; Wang, Y.; Duan, J.; Yang, T.; Huang, C.; Tong, Y.; and Xu, B. 2022.
\newblock Ts2vec: Towards universal representation of time series.
\newblock In \emph{Proceedings of the AAAI Conference on Artificial Intelligence}, volume~36, 8980--8987.

\bibitem[{Zeng et~al.(2023)Zeng, Chen, Zhang, and Xu}]{dlinear}
Zeng, A.; Chen, M.; Zhang, L.; and Xu, Q. 2023.
\newblock Are transformers effective for time series forecasting?
\newblock In \emph{Proceedings of the AAAI conference on artificial intelligence}, volume~37, 11121--11128.

\bibitem[{Zerveas et~al.(2021)Zerveas, Jayaraman, Patel, Bhamidipaty, and Eickhoff}]{TST}
Zerveas, G.; Jayaraman, S.; Patel, D.; Bhamidipaty, A.; and Eickhoff, C. 2021.
\newblock A transformer-based framework for multivariate time series representation learning.
\newblock In \emph{Proceedings of the 27th ACM SIGKDD conference on knowledge discovery \& data mining}, 2114--2124.

\bibitem[{Zhai et~al.(2023)Zhai, Likhomanenko, Littwin, Busbridge, Ramapuram, Zhang, Gu, and Susskind}]{AttentionColl3}
Zhai, S.; Likhomanenko, T.; Littwin, E.; Busbridge, D.; Ramapuram, J.; Zhang, Y.; Gu, J.; and Susskind, J.~M. 2023.
\newblock Stabilizing transformer training by preventing attention entropy collapse.
\newblock In \emph{International Conference on Machine Learning}, 40770--40803. PMLR.

\bibitem[{Zhang, Wang, and Wang(2022)}]{maskTheo}
Zhang, Q.; Wang, Y.; and Wang, Y. 2022.
\newblock How mask matters: Towards theoretical understandings of masked autoencoders.
\newblock \emph{Advances in Neural Information Processing Systems}, 35: 27127--27139.

\bibitem[{Zhou et~al.(2021)Zhou, Zhang, Peng, Zhang, Li, Xiong, and Zhang}]{Informer}
Zhou, H.; Zhang, S.; Peng, J.; Zhang, S.; Li, J.; Xiong, H.; and Zhang, W. 2021.
\newblock Informer: Beyond efficient transformer for long sequence time-series forecasting.
\newblock In \emph{Proceedings of the AAAI conference on artificial intelligence}, volume~35, 11106--11115.

\bibitem[{Zhou et~al.(2022)Zhou, Ma, Wen, Wang, Sun, and Jin}]{fedformer}
Zhou, T.; Ma, Z.; Wen, Q.; Wang, X.; Sun, L.; and Jin, R. 2022.
\newblock Fedformer: Frequency enhanced decomposed transformer for long-term series forecasting.
\newblock In \emph{International conference on machine learning}, 27268--27286. PMLR.

\end{thebibliography}
\appendix
% \newtheorem{lemma}{Lemma}
% \newtheorem{corollary}{Corollary}
% END REMOVE bibentry

\section{Related Works}
\subsection{Time Series Forecasting}
Time series forecasting has seen significant advancements in recent years, particularly the Transformer-based models, which have proven highly successful in supervised learning. 
While earlier works \cite{Informer, autoformer, fedformer} involved modifications to the key components of the vanilla Transformer \cite{Transformer}, leading state-of-the-art methods PatchTST \cite{PatchTST}and iTransformer \cite{itransformer}implement minimal alterations. PatchTST applies patching to the input series and maintains channel independence, and iTransformer simply inverts the temporal and channel dimensions of the input. Notably, both PatchTST and iTransformer achieve state-of-the-art performance by solely altering the shape of the input time series.

In addition, time series forecasting has incorporated numerous self-supervised learning techniques, which have already demonstrated substantial progress in natural language processing (NLP)\cite{gpt,BERT,raffel2020exploring,gpt3} and computer vision (CV)\cite{moco, simmim, BeiT,MaskedAE}. 
Self-supervised learning aims to extract knowledge from large-scale, multi-domain unlabeled data, yielding valuable and generalizable representations.
These techniques mainly include contrastive learning and masked modeling. Compared to contrastive learning methods, masked modeling tends to perform better in time series forecasting tasks because it can capture more low-level information according to \cite{xie2023revealing, simmim}. 
Mask modeling can further enhance performance under in-domain forecasting scenarios \cite{PatchTST}, where models are pre-trained and fine-tuned on the same dataset. This improvement aligns with the latest results presented in \cite{nevertrainfrom}. In the meanwhile, mask modeling also shows the promising results in cross-domaim forecasting scenarios\cite{PatchTST,SimMTM}.

\subsection{Masked Time Series Modeling}
Masked modeling is an essential pre-training technique that trains models by reconstructing the masked content based on the visible information. Leveraging advancements in NLP and computer vision, masked time series modeling has become crucial in time series forecasting. 
This approach enables models to learn more robust and general representations, which are beneficial across various downstream forecasting datasets and domains. 

TST\cite{TST} first applies point-level masking strategy into time series analysis using a Transformer-based framework. TimeMAE\cite{timemae} integrates both masked codeword classification and masked representation regression to pre-train the model effectively.
SimMTM\cite{SimMTM} reconstructs the masked content by weighted aggregation of multiple masked series.
PatchTST\cite{PatchTST} employs patching technique and develops a patch-level masking strategy, which has led to significant advancements in forecasting tasks, establishing it as an ideal backbone for further time series pre-training studies. It has become a common practice in time series forecasting to segment time series into patches. This approach effectively encapsulates local dynamics within input tokens, enhancing the model's ability to capture and analyze temporal patterns.\cite{Survey1}. Building upon the patching technique, numerous time series foundation model works have emerged and achieve significant performance in time series forecasting \cite{moment,moirai}.

\section{Theoretical Analysis: slowing down the rank collapse of Transformer}

\begin{lemma}
Let $\mathrm{SAN}$ denote a self-attention layer, and consider stacking $L$ such layers. Then, under certain conditions, the representations within the stacked self-attention layers will converge to a rank-1 matrix as $L \to \infty$.
\end{lemma}

\begin{proof}
Similar with \cite{AttentionColl2}, consider the residual defined by
\begin{equation}
    \mathrm{res}(\mathbf{X}) = \mathbf{X} - \mathbf{1}\mathbf{x}^\top,
\end{equation}
where \(\mathbf{x} = \frac{1}{n}\mathbf{1}^\top\mathbf{X}\) and \(\mathbf{1}\in\mathbb{R}^n\) is the all-ones vector. When \(\mathrm{res}(\mathbf{X}) = \mathbf{0}\), the matrix \(\mathbf{X}\) has identical rows and thus is rank-1. Let \(\mathrm{SAN}\) be a self-attention layer and assume there exist constants \(\gamma,\beta,d\) such that for any \(\mathbf{X}\in\mathbb{R}^{n\times d}\),
\begin{equation}
    \|\mathrm{res}(\mathrm{SAN}(\mathbf{X}))\|_{1,\infty} \leq C \|\mathrm{res}(\mathbf{X})\|_{1,\infty}^3,
\end{equation}
where
\begin{equation}
    C = \frac{4\gamma\beta}{\sqrt{d}}.
\end{equation}
Define
\begin{equation}
    r_L = \|\mathrm{res}(\mathrm{SAN}^L(\mathbf{X}))\|_{1,\infty}, \quad r_0 = \|\mathrm{res}(\mathbf{X})\|_{1,\infty}.
\end{equation}
For \(L=1\), we have
\begin{equation}
    r_1 \leq C r_0^3.
\end{equation}
By induction, assume
\begin{equation}
    r_L \leq C^{\frac{3^L - 1}{2}} r_0^{3^L}.
\end{equation}
Applying the single-layer inequality to \(\mathrm{SAN}^L(\mathbf{X})\),
\begin{equation}
    r_{L+1} \leq C (r_L)^3.
\end{equation}
Substituting the inductive hypothesis,
\begin{equation}
    r_{L+1} \leq C\bigl(C^{\tfrac{3^L -1}{2}} r_0^{3^L}\bigr)^3 = C^{\frac{3^{L+1}-1}{2}} r_0^{3^{L+1}},
\end{equation}
so the induction is complete and the inequality holds for all \(L\). 

Next, considering the growth rate of $r_{L}$:
\begin{equation}
    \ln(r_L) \leq \frac{3^L -1}{2}\ln(C) + 3^L \ln(r_0)
\end{equation}
% As $L$ increases, the terms on the order of 
% $3^{L}$ dominate. If the initial residual $r_{0}$ and the constant $C$ satisfy the condition that $\ln{r_{0}}+\frac{1}{2}\ln{C}<0$, then the decay of $r_{0}^{3^{L}}$ outpaces the growth of $C^{\frac{3^{L}-1}{2}}$, resulting in $r_{L}\rightarrow 0$,

Since \(\ln(r_L) \leq \frac{3^L -1}{2}\ln(C) + 3^L \ln(r_0)\), the behavior as \(L\) grows large depends on the sign of \(\ln(r_0) + \tfrac{1}{2}\ln(C)\). If \(r_0 < C^{-1/2}\), then \(\ln(r_0) + \tfrac{1}{2}\ln(C)<0\), and thus the term \(r_0^{3^L}\) vanishes faster than \(C^{\frac{3^L -1}{2}}\) can grow, implying \(r_L \to 0\). Therefore, if \(r_0 < C^{-1/2}\), \(\mathrm{res}(\mathrm{SAN}^L(\mathbf{X}))\) converges to \(\mathbf{0}\) as \(L \to \infty\), and hence \(\mathrm{SAN}^L(\mathbf{X})\) converges to a rank-1 matrix.

\end{proof}

\begin{corollary}
The DropPatch strategy effectively slows down the rate at which the representation matrix of a Transformer degenerates into a rank-1 matrix.
\end{corollary}

\begin{proof}
Suppose $\mathbf{X}\in\mathbb{R}^{L\times d}$ is the input representation matrix and 

\begin{equation}
    A = \mathrm{Softmax}\bigl((\mathbf{X}W_Q+\mathbf{1}b_Q^\top)(\mathbf{X}W_K+\mathbf{1}b_K^\top)^\top\bigr)\in\mathbb{R}^{L\times L}
\end{equation}
the corresponding attention matrix. Consider a scenario where we form $\mathbf{X}'\in \mathbb{R}^{L'\times d}$ by uniformly and independently dropping rows, remaining $L'<L$ rows from $\mathbf{X}$, and letting
\begin{equation}
    A'=\mathrm{Softmax}\bigl((\mathbf{X}'W_Q+\mathbf{1}b_Q^\top)(\mathbf{X}'W_K+\mathbf{1}b_K^\top)^\top\bigr)\in\mathbb{R}^{L'\times L'}
\end{equation}
be the new attention matrix. We assume that there exist real numbers $\mu_i$ and small perturbations $\delta_{ij}$ such that for $S_{ij}=(\mathbf{X}W_Q)_i(\mathbf{X}W_K)_j^\top$, we have $S_{ij}=\mu_i+\delta_{ij}$ with $\sum_j\delta_{ij}=0$ and $|\delta_{ij}|\leq \epsilon$ for all $(i,j)$, where $\epsilon>0$ is sufficiently small. Under this assumption, we can approximate $\exp(\mu_i+\delta_{ij})=\exp(\mu_i)(1+\delta_{ij}+O(\epsilon^2))$, and the softmax denominator $\sum_{k=1}^L \exp(\mu_i+\delta_{ik})=L\exp(\mu_i)(1+O(\epsilon^2))$ via Taylor expansion. Consequently,
\begin{equation}
    \begin{aligned}
         A_{ij} = &\frac{\exp(\mu_i+\delta_{ij})}{\sum_{k}\exp(\mu_i+\delta_{ik})} \\
        &= \frac{1+\delta_{ij}+O(\epsilon^2)}{L(1+O(\epsilon^2))} \\
        &= \frac{1+\delta_{ij}}{L} + O(\epsilon^2).
    \end{aligned}
\end{equation}

Similarly,
\begin{equation}
    A'_{ij}=\frac{1+\delta_{ij}}{L'}+O(\epsilon^2).
\end{equation}

Define $\Delta_i=\max_{j,j'}|\delta_{ij}-\delta_{ij'}|$. Then
\begin{equation}
    |A_{ij}-A_{ij'}|=\frac{|\delta_{ij}-\delta_{ij'}|+O(\epsilon^2)}{L}\leq \frac{\Delta_i}{L}+O(\epsilon^2),
\end{equation}
\begin{equation}
    |A'_{ij}-A'_{ij'}|\leq \frac{\Delta_i}{L'}+O(\epsilon^2).
\end{equation}
Thus
\begin{equation}
    \max_{j,j'}|A'_{ij}-A'_{ij'}|=\frac{\Delta_i}{L'}+O(\epsilon^2),
\end{equation}
and
\begin{equation}
    \max_{j,j'}|A_{ij}-A_{ij'}|=\frac{\Delta_i}{L}+O(\epsilon^2).
\end{equation}
As $\epsilon\to 0$, we get
\begin{equation}
    \frac{\max_{j,j'}|A'_{ij}-A'_{ij'}|}{\max_{j,j'}|A_{ij}-A_{ij'}|}\approx \frac{L}{L'}>1.
\end{equation}

Next, consider $\sum_{i=1}^L \max_{j,j'}|A_{ij}-A_{ij'}|\approx \frac{1}{L}\sum_{i=1}^L \Delta_i + O(\epsilon^2)$. After row dropping, let $\mathcal{I}'$ be the set of remained $L'$ rows. Then
\begin{equation}
    \sum_{i=1}^{L'}\max_{j,j'}|A'_{ij}-A'_{ij'}|\approx \frac{1}{L'}\sum_{i\in \mathcal{I}'}\Delta_i + O(\epsilon^2).
\end{equation}
Since each row is chosen with probability $p=L'/L$, the expectation satisfies
\begin{equation}
    \mathbb{E}\left[\frac{1}{L'}\sum_{i\in \mathcal{I}'} \Delta_i\right]=\frac{1}{L'}(pL)\overline{\Delta}=\overline{\Delta},
\end{equation}
where $\overline{\Delta}=\frac{1}{L}\sum_{i=1}^L \Delta_i$. Hence in expectation and with high probability (using concentration inequalities if $\Delta_i$ are bounded and weakly dependent), we have
\begin{equation}
    \mathbb{E}\left[\sum_{i=1}^{L'}\max_{j,j'}|A'_{ij}-A'_{ij'}|\right]\approx \frac{1}{L}\sum_{i=1}^L \Delta_i + O(\epsilon^2),
\end{equation}
showing that $\sum_i\max_{j,j'}|A_{ij}-A_{ij'}|$ remains essentially unchanged by row dropping.

Now consider $\max_{j,j'}\sum_i|A_{ij}-A_{ij'}|$. Since $\sum_j\delta_{ij}=0$, we get
\begin{equation}
    \sum_{i=1}^L|A_{ij}-A_{ij'}|\approx L|\overline{A}_j-\overline{A}_{j'}|+O(\epsilon^2).
\end{equation}
After row dropping,
\begin{equation}
    \sum_{i=1}^{L'}|A'_{ij}-A'_{ij'}|\approx L'|\overline{A}_j-\overline{A}_{j'}|+O(\epsilon^2),
\end{equation}
hence
\begin{equation}
    \frac{\sum_{i=1}^{L'}|A'_{ij}-A'_{ij'}|}{\sum_{i=1}^L|A_{ij}-A_{ij'}|}\approx \frac{L'}{L}<1.
\end{equation}

Recall \cite{AttentionColl2}, we have:
\begin{equation}
    \gamma \geq \frac{\sqrt{\max_{i,j,j'}|A_{ij}-A_{ij'}|\sum_i\max_{j,j'}|A_{ij}-A_{ij'}|}}{\max_{j,j'}\sum_i|A_{ij}-A_{ij'}|},
\end{equation}
the numerator is influenced by a factor that increases approximately by $L/L'$, while the denominator decreases by about $L'/L$. Thus after the row dropping, the new $\gamma'$ satisfies an inequality with a larger lower bound, roughly scaling as
\begin{equation}
    \gamma' \geq \gamma \cdot \frac{\sqrt{L/L'}}{(L'/L)}=\gamma \cdot \frac{L}{L'}\sqrt{\frac{L}{L'}}> \gamma.
\end{equation}

According to \cite{AttentionColl2}, we have an inequality of the form
\begin{equation}
    r_{L+1}\leq\left(\frac{4\gamma\beta}{\sqrt{d}}\right)^{\frac{3^L-1}{2}}r_L^{3^L},
\end{equation}
implying
\begin{equation}
    \frac{r_{L+1}}{r_L}\leq \left(\frac{4\gamma\beta}{\sqrt{d}}\right)^{\frac{3^L-1}{2}}r_L^{3^L-1}.
\end{equation}
Since $\gamma'$ is larger than $\gamma$, the upper bound on $\frac{r_{L+1}}{r_L}$ increases, causing the residual to shrink more slowly layer by layer and thus delaying the rank-1 degeneration of the representation matrix. Under the assumptions of sufficiently small $\epsilon$, independent uniform random selection of rows, and bounded (or weakly correlated) $\Delta_i$, the argument holds in expectation and with high probability. Therefore, the DropPatch operation effectively slows down the rate at which the representation matrix degenerates to a rank-1 matrix.

\end{proof}

\section{Implementation Details}
Experiments are conducted five times, implemented using Pytorch \cite{pytorch}, and carried out on a single NVIDIA Tesla V100-SXM2-32GB GPU. We replicate the baseline methods based on their official implementations and adhere to the configurations specified in their original papers. We utilize the mean square error (MSE) and mean absolute error (MAE) for the time series forecasting.

\subsection{Datasets}
\label{sec:dataset}

Table \ref{tab:datasets} presents the information about the 12 public datasets used in experiments. The ETT datasets\cite{Informer} track various electrical transformer statistics such as load capacity and oil temperature. The ECL dataset\cite{autoformer} records electricity consumption for 321 clients from 2012 to 2014, measured in kilowatts. The Traffic dataset\cite{autoformer} monitors road occupancy rates using data from 862 sensors located on freeways in the San Francisco Bay area, encompassing 48 months of hourly data from 2015 to 2016, provided by the California Department of Transportation. The Weather dataset\cite{autoformer} comprises 21 meteorological factors collected every 10 minutes throughout 2020 from the Weather Station at the Max Planck Biogeochemistry Institute. The Exchange dataset\cite{autoformer} tracks the daily exchange rates of eight currencies (Australian Dollar, Pound Sterling, Canadian Dollar, Swiss Franc, Chinese Yuan, Japanese Yen, New Zealand Dollar, and Singapore Dollar) against the US Dollar, covering a span of 26 years from 1990 to 2016. The PEMS dataset includes public traffic network data from California, collected in 5-minute intervals. For our experiments, we utilize the same four public subsets (PEMS03, PEMS04, PEMS07, PEMS08) adopted in \cite{itransformer,scinet}.

For STS66M dataset, we merged 10 datasets from different domains, with various frequency. The details of the datasets are presented in Table \ref{tab:sts}. We also add ECL and PEMS07 to STS66M to formulate a larger synthesized dataset, namely STS162M. Details about ECL and PEMS07 can be found in Table \ref{tab:datasets}. STS66M contains over 3.76 million data points, with a file size 66M in total; STS162M contains over 32.5 million data points, with a file size 162M in total. 

As the different datasets contain various number of variants, we split all the datasets to univariate series for a convenient pre-training. When conducting pre-training on STS datasets, the batch size is set to 8192. 

\begin{table*}[h]
  \caption{Dataset description in detail. \emph{Feature} denotes the number of variates; \emph{Size} denotes the total number of time points in Train/valid/test set; \emph{Frequency} denotes the sampling rate. \emph{Usage} denotes in which experiments the datasets are used.}
  \label{tab:datasets}
  \centering
  \scriptsize
  \begin{tabular}{l|c|c|c|c|c}
    \toprule
    Datasets&Feature&Size&Frequency&Task&Information\\
    \midrule
    ETTh1 \& ETTh2 & 7 & (8545,2881,2881) & Hourly & In \& Cross-domain, Few-shot & Device\\
    ETTm1 \& ETTm2 & 7 & (34465,11521,11521) & 15min & In \&  Cross-domain, Few-shot & Device\\
    Weather& 21 & (36792,5271,10540) & 10min & In \& Cross-domain, Few-shot & Weather\\
    ECL& 321 & (18317,2633,5261) & Hourly & In \& Cross-domain, Few-shot & Electricity\\
    Traffic& 862 & (12185,1757,3509) & Hourly & In \& Cross-domain, Few-shot & Transportation\\
    Exchange& 8 & (5120,665,1422) & Daily & Cold-start & Economy\\
    PEMS03& 358 & (15617,5135,5135) & 5min & Cold-start & Transportation\\
    PEMS04& 307 & (10172,3375,281) & 5min & Cold-start & Transportation\\
    PEMS07& 883 & (16911,5622,468) & 5min & Cold-start & Transportation\\
    PEMS08& 170 & (10690,3548,265) & 5min & Cold-start & Transportation\\
  \bottomrule
\end{tabular}
\end{table*}

\begin{table*}[h]
  \caption{STS dataset description in detail. \emph{Feature} denotes the number of variates; \emph{Size} denotes the total number of time points; \emph{Frequency} denotes the sampling rate. STS dataset comprises over 3.76 million data points in total. We partitioned the entire dataset into training and validation sets with ratios of 0.8 and 0.2, respectively. The STS dataset is specifically merged for pre-training purposes, thus does not include a testing set.}
  \label{tab:sts}
  \centering
  \setlength{\tabcolsep}{2.2pt}
  \scriptsize
  \begin{tabular}{lc|c|c|c|c|c}
    \toprule
    Datasets&&Feature&Size&Frequency&Information&Source\\
    \midrule
     Wind Power& & 1 & 493144 & 1min & Energy & Monash Time Series Forecasting Archive\cite{monash}\\
     Solor Power& & 1 & 493149 & 1min & Energy & Monash Time Series Forecasting Archive\cite{monash}\\
     Sunspot& & 1 & 73924 & Daily & Nature & Monash Time Series Forecasting Archive\cite{monash}\\
     Saugeen River Flow& & 1 & 23741 & Daily & Nature & Monash Time Series Forecasting Archive\cite{monash}\\
     Aus. Electricity Demand& &5 & 230736 & 30min & Electricity & Monash Time Series Forecasting Archive\cite{monash}\\
     Appliances Energy& & 26 & 19735 & 10min & Energy & UCIMachine Learning Repository\cite{appliance_energy, applyenergy}\\
     Metro Interstate Traffic Volume& & 1 & 48204 & Hourly & Traffic & UCIMachine Learning Repository\cite{metrovolume}\\
     Power Consumption of Tetouan City&& 8 & 52417 & 10min & Social & UCIMachine Learning Repository\cite{pctc}\\
     Air Quality& & 13 & 9358 & 5min & Air Quality & UCIMachine Learning Repository\cite{air_quality}\\
     USWeather& & 12 & 35064 & Hourly & Weather & \url{https://www.ncei.noaa.gov/data/local-climatological-data/}{USWeather}\\
  \bottomrule
\end{tabular}
\end{table*}

The Wind Power dataset comprises a lengthy daily time series detailing wind power production in megawatts (MW), with measurements taken every 4 seconds starting from August 1, 2019. We subsample this dataset to a one-minute frequency for analysis. Similarly, the Solar Power dataset records solar power production in MW every 4 seconds from the same start date, which we also subsample to one-minute intervals.
The Sunspot dataset includes a historical daily time series of sunspot numbers, spanning from January 8, 1818, to May 31, 2020. This dataset provides a detailed record of solar activity over more than two centuries.
The Saugreen River Flow dataset documents the daily mean flow of the Saugeen River at Walkerton in cubic meters per second, covering a period from January 1, 1915, to December 31, 1979. This dataset is valuable for studying long-term changes in river flow.
The Aus. Electricity Demand dataset contains five time series, each representing the half-hourly electricity demand of five Australian states: Victoria, New South Wales, Queensland, Tasmania, and South Australia.
These datasets are accessible as part of the collection described in \cite{monash}.

We also collect data from the UCI Machine Learning Repository. The Appliances Energy dataset \cite{appliance_energy,applyenergy} logs energy usage data every 10 minutes over approximately 4.5 months. It includes two main types of data: energy consumption recorded every 10 minutes using m-bus energy meters and weather data from the nearest airport weather station (Chievres Airport, Belgium). The weather data, obtained from a public dataset from Reliable Prognosis (rp5.ru), is merged with the energy data using the date and time rows.
The Metro Interstate Traffic Volume dataset \cite{metrovolume} provides hourly traffic volume data for westbound I-94 in Minneapolis-St. Paul, MN, from 2012 to 2018, including weather conditions.
The Power Consumption of Tetouan City dataset \cite{pctc} pertains to the electricity consumption of three different distribution networks in Tetouan city, located in northern Morocco.
The Air Quality dataset \cite{air_quality} features hourly readings from a gas multisensor device deployed in an Italian city, capturing various air quality indicators.
Additionally, we include the USWeather dataset in our synthesized dataset, as introduced by \cite{Informer}.

\subsection{Experiments Implementation}
\label{sec:implementation}
\subsubsection{Baseline Setup.}
For PatchTST, we rely on the officially reported results from the original study \cite{PatchTST}, which is a strong baseline.
The official implementation for SimMTM uses a lookback length of 336. However, we have observed a degradation in SimMTM's performance when increasing the pre-training look-back length to 512 on ETT and Weather datasets. And an Out-of-Memory (OOM) issue arises when pre-training on the ECL and Traffic datasets, even with a batch size of 1. Therefore, we directly report results from the official paper \cite{SimMTM}.

For other baseline models, experiments are conducted using the official code and configurations at two different lookback lengths, 336 and 512. We choose and report the better results from these tests.

\subsubsection{Pre-training Setup.} 
DropPatch is pre-trained for 50 epochs using a learning rate of $1e-3$, and batch sizes are set at either 16 or 64. Specifically, for pre-training on STS datasets, the batch size is significantly increased to 8192 to accommodate the extensive dataset size. Across all experiments, the drop ratio is consistently fixed at 0.6 and the mask ratio at 0.4. The OneCycle learning rate schedule is utilized to optimize training dynamics.
We fix the lookback length to 512 for our method following the self-supervised PatchTST\cite{PatchTST}. 

\subsubsection{Fine-tuning Setup.} We fine-tune DropPatch for 1 epoch under in-domain and full fine-tuning settings. The learning rate is in $\{0.001, 0.0003, 0.0001, 0.00003, 0.00001\}$ and the batch size is in $\{4, 8, 16\}$. For cold-start and few-shot scenarios, we fine-tune the models for 10 epochs. The learning rate is in $\{0.001, 0.0001\}$ and the batch size is in $\{8, 16, 32, 64\}$. 

\subsubsection{Model Parameters.} DropPatch is typically configured with three encoder layers, each featuring 16 attention heads ($H = 16$) and a latent space dimension of 128 ($D = 128$). The feed-forward network within these encoders has a dimension of $F = 256$ and uses the GELU activation function\cite{gelu}. For smaller datasets like ETTh1 and ETTh2, particularly under in-domain settings, we adapt the model configuration to a smaller encoder with three layers, but with fewer attention heads ($H = 4$), a reduced latent space dimension ($D=16$), and a smaller feed-forward dimension ($F=128$). When conduct fully fine-tuning on larger datasets such as Traffic and ECL, we utilize a larger encoder configuration with four layers, maintaining $H=16$ and $F=256$ but increasing the latent space dimension to 256 ($D=256$). 

\begin{figure*}[ht]
  \centering
  \includegraphics[width=\textwidth]{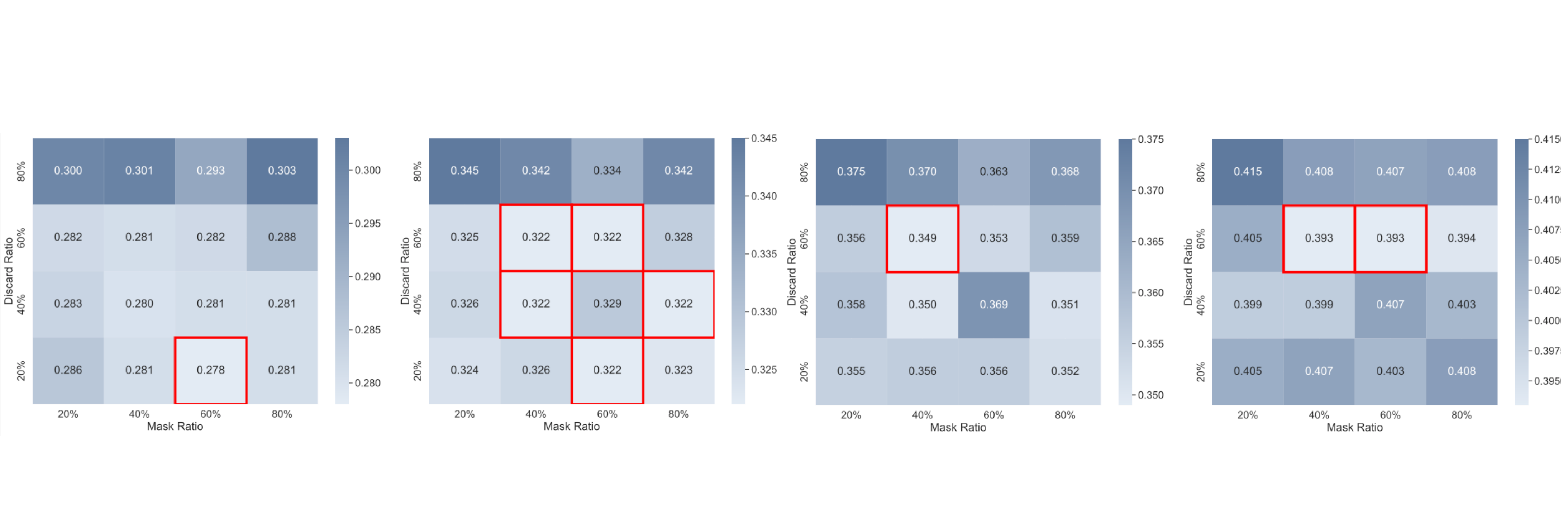}
  \caption{MSE performance of DropPatch on ETTm1 dataset (in-domain setting) with different dropping ratio and masking ratio. The forecasting step from left to right is 96 192, 336, 720. The standard deviations of DropPatch are within 0.002 for MSE and within 0.001 for MAE.}
  \label{fig:drop_mask_full}
\end{figure*}

\section{Full Results}
\label{sec:full_results}
We present the full comparison results of all experimental performances in this section.

\subsection{Full Results of In-domain Forecasting}
\label{sec:full_results_indomain}
The comparison of full results of the four forecasting steps across all 7 datasets with the 7 baseline in-domain tasks is shown in Table 
\ref{tab:in-domain_big}.

\begin{table*}[h]
\setlength{\tabcolsep}{2.5pt}
\caption{In-domain time series forecasting full results, forecasting steps $T \in \{96, 192, 336, 720\}$. Models are pre-trained and fine-tuned in the same dataset. For ETTh1, the standard deviations of DropPatch are within 0.005 for MSE and within 0.004 for MAE. For other datasets, the standard deviations of DropPatch are within 0.002 for MSE and within 0.001 for MAE.}
\label{tab:in-domain_big}
\centering
\scriptsize
\begin{tabular}{lc|c|c|c|c|c|c|c|c}
\toprule
& Models & DropPatch & PatchTST & SimMTM & Ti-MAE & TST & LaST & CoST & TS2Vec\\

& Metrics & MSE ~ MAE & MSE ~ MAE & MSE ~ MAE & MSE ~ MAE & MSE ~ MAE & MSE ~ MAE & MSE ~ MAE & MSE ~ MAE \\
\midrule
\multirow{5}{*}{ETTh1} 
& 96 & 0.369 ~ 0.406 & \textbf{0.366 ~ 0.397} & 0.367 ~ 0.402 & 0.708 ~ 0.570 & 0.503 ~ 0.527 & 0.399 ~ 0.412 & 0.514 ~ 0.512 & 0.493 ~ 0.511 \\
& 192 & \textbf{0.402} ~ 0.426 & 0.431 ~ 0.443 & 0.403 ~ \textbf{0.425} & 0.725 ~ 0.587 & 0.601 ~ 0.552 & 0.484 ~ 0.468 & 0.655 ~ 0.590 & 0.617 ~ 0.732 \\ 
& 336 & \textbf{0.409} ~ 0.433 & 0.450 ~ 0.456 & 0.415 ~ \textbf{0.430} & 0.713 ~ 0.589 & 0.625 ~ 0.541 & 0.580 ~ 0.533 & 0.790 ~ 0.666 & 0.818 ~ 0.807 \\ 
& 720 & \textbf{0.419 ~ 0.452} & 0.472 ~ 0.484 & 0.430 ~ 0.453 & 0.736 ~ 0.618 & 0.768 ~ 0.628 & 0.432 ~ 0.432 & 0.880 ~ 0.739 & 1.190 ~ 0.863  \\
& AVG & \textbf{0.400} ~ 0.429 & 0.430 ~ 0.445 & 0.404 ~ \textbf{0.428} & 0.721 ~ 0.591 & 0.624 ~ 0.562 & 0.474 ~ 0.461 & 0.710 ~ 0.627 & 0.643 ~ 0.728  \\
\midrule 
\multirow{5}{*}{ETTh2} 
& 96 & \textbf{0.275 ~ 0.339} & 0.284 ~ 0.343 & 0.288 ~ 0.347 & 0.443 ~ 0.465 & 0.335 ~ 0.392 & 0.331 ~ 0.390 & 1.061 ~ 0.819 & 0.541 ~ 0.673  \\
& 192 & \textbf{0.343 ~ 0.382} & 0.355 ~ 0.387 & 0.346 ~ 0.385 & 0.533 ~ 0.516 & 0.444 ~ 0.441 & 0.751 ~ 0.612 & 1.669 ~ 0.998 & 0.680 ~ 0.712  \\ 
& 336 & 0.367 ~ 0.402 & 0.379 ~ 0.411 & \textbf{0.363 ~ 0.401} & 0.445 ~ 0.472 & 0.455 ~ 0.494 & 0.460 ~ 0.478 & 1.856 ~ 1.052 & 0.753 ~ 0.882  \\ 
& 720 & 0.403 ~ 0.437 & 0.400 ~ 0.435 & \textbf{0.396 ~ 0.431} & 0.507 ~ 0.498 & 0.481 ~ 0.504 & 0.552 ~ 0.509 & 2.049 ~ 1.097 & 1.231 ~ 1.156  \\
& AVG & \textbf{0.347 ~ 0.390} & 0.355 ~ 0.394 & 0.348 ~ 0.391 & 0.482 ~ 0.488 & 0.429 ~ 0.458 & 0.499 ~ 0.497 & 1.659 ~ 0.992 & 0.801 ~ 0.856  \\
\midrule 
\multirow{5}{*}{ETTm1}
& 96 & \textbf{0.281 ~ 0.338} & 0.289 ~ 0.344 & 0.289 ~ 0.343 & 0.647 ~ 0.497 & 0.454 ~ 0.456 & 0.322 ~ 0.361 & 0.376 ~ 0.420 & 0.563 ~ 0.551 \\
& 192 & \textbf{0.322 ~ 0.367} & 0.323 ~ 0.368 & 0.323 ~ 0.369 & 0.597 ~ 0.508 & 0.471 ~ 0.490 & 0.348 ~ 0.373 & 0.420 ~ 0.451 & 0.599 ~ 0.558 \\ 
& 336 & \textbf{0.349} ~ 0.388 & 0.353 ~ 0.387 & 0.349 ~ \textbf{0.385} & 0.699 ~ 0.525 & 0.457 ~ 0.451 & 0.392 ~ 0.409 & 0.482 ~ 0.494 & 0.685 ~ 0.594 \\ 
& 720 & \textbf{0.393 ~ 0.418} & 0.398 ~ 0.416 & 0.399 ~ 0.418 & 0.786 ~ 0.596 & 0.594 ~ 0.488 & 0.471 ~ 0.451 & 0.628 ~ 0.578 & 0.831 ~ 0.698 \\
& AVG & \textbf{0.336 ~ 0.378} & 0.341 ~ 0.379 & 0.340 ~ 0.379 & 0.682 ~ 0.532 & 0.494 ~ 0.471 & 0.383 ~ 0.399 & 0.477 ~ 0.486 & 0.669 ~ 0.600 \\
\midrule 
\multirow{5}{*}{ETTm2}
& 96 & 0.167 ~ 0.259 & \textbf{0.166 ~ 0.256} & 0.166 ~ 0.257 & 0.304 ~ 0.357 & 0.363 ~ 0.301 & 0.160 ~ 0.254 & 0.327 ~ 0.418 & 0.275 ~ 0.353 \\
& 192 & 0.223 ~ \textbf{0.295} & \textbf{0.221 ~ 0.295} & 0.223 ~ \textbf{0.295} & 0.334 ~ 0.387 & 0.342 ~ 0.364 & 0.225 ~ 0.300 & 0.537 ~ 0.554 & 0.313 ~ 0.351 \\ 
& 336 & \textbf{0.271 ~ 0.326} & 0.278 ~ 0.333 & 0.282 ~ 0.334 & 0.420 ~ 0.441 & 0.414 ~ 0.361 & 0.239 ~ 0.366 & 0.824 ~ 0.705 & 0.352 ~ 0.387 \\ 
& 720 & \textbf{0.355 ~ 0.381} & 0.365 ~ 0.388 & 0.370 ~ 0.385 & 0.508 ~ 0.481 & 0.580 ~ 0.456 & 0.397 ~ 0.382 & 1.492 ~ 0.948 & 0.496 ~ 0.587 \\
& AVG & \textbf{0.254 ~ 0.315} & 0.258 ~ 0.318 & 0.260 ~ 0.318 & 0.392 ~ 0.417 & 0.425 ~ 0.371 & 0.255 ~ 0.326 & 0.795 ~ 0.656 & 0.359 ~ 0.420 \\
\midrule 
\multirow{5}{*}{Weather}
& 96 & \textbf{0.142 ~ 0.191} & 0.144 ~ 0.193 & 0.151 ~ 0.202 & 0.216 ~ 0.280 & 0.292 ~ 0.370 & 0.153 ~ 0.211 & 0.797 ~ 0.646 & 0.231 ~ 0.285 \\
& 192 & \textbf{0.188 ~ 0.236} & 0.190 ~ 0.236 & 0.223 ~ 0.295 & 0.303 ~ 0.335 & 0.410 ~ 0.473 & 0.207 ~ 0.250 & 0.794 ~ 0.667 & 0.393 ~ 0.412 \\ 
& 336 & \textbf{0.239 ~ 0.277} & 0.244 ~ 0.280 & 0.246 ~ 0.283 & 0.351 ~ 0.358 & 0.434 ~ 0.427 & 0.249 ~ 0.264 & 1.029 ~ 0.771 & 0.771 ~ 0.893 \\ 
& 720 & \textbf{0.312 ~ 0.330} & 0.320 ~ 0.335 & 0.320 ~ 0.338 & 0.425 ~ 0.399 & 0.539 ~ 0.523 & 0.319 ~ 0.320 & 1.361 ~ 0.916 & 1.235 ~ 1.412 \\
& AVG & \textbf{0.220 ~ 0.259} & 0.225 ~ 0.261 & 0.235 ~ 0.280 & 0.324 ~ 0.343 & 0.419 ~ 0.448 & 0.232 ~ 0.261 & 0.995 ~ 0.750 & 0.658 ~ 0.751 \\
\midrule 
\multirow{5}{*}{ECL}
& 96 & \textbf{0.126 ~ 0.217} & \textbf{0.126} ~ 0.221 & 0.133 ~ 0.223 & 0.399 ~ 0.412 & 0.292 ~ 0.370 & 0.166 ~ 0.254 & 0.230 ~ 0.353 & 0.322 ~ 0.401 \\
& 192 & \textbf{0.145 ~ 0.236} & \textbf{0.145} ~ 0.238 & 0.147 ~ 0.237 & 0.400 ~ 0.460 & 0.270 ~ 0.373 & 0.178 ~ 0.278 & 0.253 ~ 0.371 & 0.343 ~ 0.416 \\ 
& 336 & \textbf{0.161 ~ 0.255} & 0.164 ~ 0.256 & 0.166 ~ 0.265 & 0.564 ~ 0.573 & 0.334 ~ 0.323 & 0.186 ~ 0.275 & 0.197 ~ 0.287 & 0.362 ~ 0.435 \\ 
& 720 & 0.196 ~ \textbf{0.288} & \textbf{0.193} ~ 0.291 & 0.203 ~ 0.297 & 0.880 ~ 0.770 & 0.344 ~ 0.346 & 0.213 ~ 0.288 & 0.230 ~ 0.328 & 0.388 ~ 0.456 \\
& AVG & \textbf{0.157 ~ 0.249} & \textbf{0.157} ~ 0.252 & 0.162 ~ 0.256 & 0.561 ~ 0.554 & 0.310 ~ 0.353 & 0.186 ~ 0.274 & 0.228 ~ 0.335 & 0.354 ~ 0.427 \\
\midrule 
\multirow{5}{*}{Traffic}
& 96 & \textbf{0.347 ~ 0.242} & 0.352 ~ 0.244 & 0.368 ~ 0.262 & 0.781 ~ 0.431 & 0.559 ~ 0.454 & 0.706 ~ 0.385 & 0.751 ~ 0.431 & 0.466 ~ 0.367 \\
& 192 & \textbf{0.368 ~ 0.250} & 0.371 ~ 0.253 & 0.373 ~ 0.251 & 0.911 ~ 0.428 & 0.583 ~ 0.493 & 0.709 ~ 0.388 & 0.751 ~ 0.424 & 0.476 ~ 0.367 \\ 
& 336 & \textbf{0.377 ~ 0.256} & 0.381 ~ 0.257 & 0.395 ~ 0.254 & 0.911 ~ 0.502 & 0.637 ~ 0.469 & 0.714 ~ 0.394 & 0.761 ~ 0.425 & 0.499 ~ 0.376 \\ 
& 720 & \textbf{0.420 ~ 0.280} & 0.425 ~ 0.282 & 0.432 ~ 0.290 & 1.106 ~ 0.530 & 0.663 ~ 0.594 & 0.723 ~ 0.421 & 0.780 ~ 0.433 & 0.563 ~ 0.390 \\
& AVG & \textbf{0.378 ~ 0.257} & 0.382 ~ 0.259 & 0.392 ~ 0.264 & 0.916 ~ 0.423 & 0.611 ~ 0.503 & 0.713 ~ 0.397 & 0.760 ~ 0.428 & 0.501 ~ 0.375 \\
\bottomrule
\end{tabular}
\end{table*}

\subsection{Full Results of Cross-domain Forecasting}
\label{sec:full_results_crossdomain}
For the evaluation of cross-domain tasks, we set up two groups of experiments: one with a fixed source dataset transferring to different target datasets, and the other using different source datasets transferring to a fixed target dataset. 
The former is shown in Table 
\ref{tab:cross-domain_big_ECL}, the latter is shown in Table \ref{tab:cross-domain_big}. 

% It is important to note that for the former, we compared several supervised learning methods since their results are directly given. For the latter, we compared all methods that are representation learning methods.

\begin{table*}[h]
\setlength{\tabcolsep}{2.5pt}
\caption{Cross-domain time series forecasting full results. The source dataset is fixed as ECL. Forecasting steps $T \in \{96, 192, 336, 720\}$. iTransformer, DLinear, and FEDformer are models that directly conduct supervised learning on the target dataset. The standard deviations of DropPatch are within 0.002 for MSE and within 0.001 for MAE.}
\label{tab:cross-domain_big_ECL}
\centering
\scriptsize
\begin{tabular}{lc|c|c|c|c|c}
\toprule
&Models & DropPatch & PatchTST & iTransformer & DLinear & FEDformer \\

&Metrics & MSE ~ MAE & MSE ~ MAE & MSE ~ MAE & MSE ~ MAE & MSE ~ MAE \\
\midrule

\multirow{5}{*}{ECL$\rightarrow$ETTm1}
& 96 & \textbf{0.287 ~ 0.342} & 0.288 ~ 0.345 & 0.311 ~ 0.366 & 0.299 ~ 0.343 & 0.326 ~ 0.390 \\
& 192 & 0.331 ~ 0.371 & \textbf{0.330} ~ 0.372 & 0.348 ~ 0.385 & 0.335 ~ \textbf{0.365} & 0.365 ~ 0.415 \\ 
& 336 & 0.364 ~ 0.393 & \textbf{0.359} ~ 0.392 & 0.380 ~ 0.405 & 0.369 ~ \textbf{0.386} & 0.392 ~ 0.425 \\ 
& 720 & 0.412 ~ 0.425 & \textbf{0.406 ~ 0.421} & 0.443 ~ 0.444 & 0.425 ~ \textbf{0.421} & 0.446 ~ 0.458 \\
& AVG & 0.349 ~ \textbf{0.383} & \textbf{0.346 ~ 0.383} & 0.371 ~ 0.400 & 0.357 ~ 0.379 & 0.382 ~ 0.422 \\
\midrule 
\multirow{5}{*}{ECL$\rightarrow$ETTm2}
& 96 & 0.169 ~ 0.262 & \textbf{0.164 ~ 0.256} & 0.179 ~ 0.273 & 0.167 ~ 0.260 & 0.180 ~ 0.271 \\
& 192 & 0.225 ~ 0.299 & \textbf{0.223 ~ 0.296} & 0.242 ~ 0.315 & 0.224 ~ 0.303 & 0.252 ~ 0.318 \\ 
& 336 & 0.281 ~ 0.338 & \textbf{0.277 ~ 0.332} & 0.291 ~ 0.345 & 0.281 ~ 0.342 & 0.324 ~ 0.364 \\ 
& 720 & \textbf{0.356 ~ 0.384} & 0.365 ~ 0.387 & 0.377 ~ 0.398 & 0.397 ~ 0.421 & 0.410 ~ 0.420 \\
& AVG & 0.258 ~ 0.321 & \textbf{0.257 ~ 0.318} & 0.272 ~ 0.333 & 0.267 ~ 0.332 & 0.292 ~ 0.343 \\
\midrule 
\multirow{5}{*}{ECL$\rightarrow$ETTh1}
& 96 & \textbf{0.365 ~ 0.397} & 0.368 ~ 0.398 & 0.400 ~ 0.425 & 0.375 ~ 0.399 & 0.376 ~ 0.415 \\
& 192 & \textbf{0.387 ~ 0.416} & 0.425 ~ 0.439 & 0.427 ~ 0.443 & 0.405 ~ 0.416 & 0.423 ~ 0.446 \\ 
& 336 & \textbf{0.396 ~ 0.428} & 0.470 ~ 0.471 & 0.454 ~ 0.464 & 0.439 ~ 0.443 & 0.444 ~ 0.462 \\ 
& 720 & \textbf{0.431 ~ 0.461} & 0.472 ~ 0.484 & 0.521 ~ 0.516 & 0.472 ~ 0.490 & 0.469 ~ 0.492 \\
& AVG & \textbf{0.395 ~ 0.426} & 0.434 ~ 0.448 & 0.451 ~ 0.462 & 0.423 ~ 0.437 & 0.428 ~ 0.454 \\
\midrule 
\multirow{5}{*}{ECL$\rightarrow$ETTh2}
& 96 & \textbf{0.282 ~ 0.344} & 0.285 ~ 0.345 & 0.299 ~ 0.359 & 0.289 ~ 0.353 & 0.332 ~ 0.374 \\
& 192 & \textbf{0.349 ~ 0.383} & 0.350 ~ 0.388 & 0.377 ~ 0.406 & 0.383 ~ 0.418 & 0.407 ~ 0.446 \\ 
& 336 & \textbf{0.365 ~ 0.403} & 0.378 ~ 0.410 & 0.429 ~ 0.442 & 0.448 ~ 0.465 & 0.400 ~ 0.447 \\ 
& 720 & 0.404 ~ \textbf{0.438} & \textbf{0.401 ~ 0.438} & 0.444 ~ 0.466 & 0.605 ~ 0.551 & 0.412 ~ 0.469 \\
& AVG & \textbf{0.350 ~ 0.392} & 0.354 ~ 0.395 & 0.387 ~ 0.418 & 0.431 ~ 0.447 & 0.388 ~ 0.434 \\
\midrule 
\multirow{5}{*}{ECL$\rightarrow$Weather}
& 96 & \textbf{0.145 ~ 0.195} & 0.145 ~ 0.195 & 0.168 ~ 0.220 & 0.176 ~ 0.237 & 0.238 ~ 0.314 \\
& 192 & \textbf{0.188 ~ 0.237} & 0.193 ~ 0.243 & 0.209 ~ 0.254 & 0.220 ~ 0.282 & 0.275 ~ 0.329 \\ 
& 336 & \textbf{0.240 ~ 0.278} & 0.244 ~ 0.280 & 0.266 ~ 0.295 & 0.265 ~ 0.319 & 0.339 ~ 0.377 \\ 
& 720 & \textbf{0.315 ~ 0.331} & 0.321 ~ 0.337 & 0.341 ~ 0.345 & 0.323 ~ 0.362 & 0.389 ~ 0.409 \\
& AVG & \textbf{0.222 ~ 0.260} & 0.226 ~ 0.264 & 0.246 ~ 0.279 & 0.246 ~ 0.300 & 0.310 ~ 0.357 \\
\midrule 
\multirow{5}{*}{ECL$\rightarrow$Traffic}
& 96 & \textbf{0.348 ~ 0.242} & 0.388 ~ 0.273 & 0.352 ~ 0.257 & 0.410 ~ 0.282 & 0.576 ~ 0.359 \\
& 192 & \textbf{0.369 ~ 0.251} & 0.400 ~ 0.277 & 0.374 ~ 0.268 & 0.423 ~ 0.287 & 0.610 ~ 0.380 \\ 
& 336 & \textbf{0.379 ~ 0.256} & 0.408 ~ 0.280 & 0.386 ~ 0.274 & 0.436 ~ 0.296 & 0.608 ~ 0.375 \\ 
& 720 & \textbf{0.419 ~ 0.278} & 0.447 ~ 0.310 & 0.409 ~ 0.284 & 0.466 ~ 0.315 & 0.621 ~ 0.375 \\
& AVG & \textbf{0.379 ~ 0.257} & 0.411 ~ 0.285 & 0.380 ~ 0.271 & 0.434 ~ 0.295 & 0.604 ~ 0.372 \\
\bottomrule
\end{tabular}
\end{table*}

\begin{table*}[h]
\setlength{\tabcolsep}{2.5pt}
\caption{Cross-domain time series forecasting full results. The target dataset is fixed as ETTm1. Forecasting steps $T \in \{96, 192, 336, 720\}$. The standard deviations of DropPatch are within 0.002 for MSE and within 0.001 for MAE.}
\label{tab:cross-domain_big}
\centering
\scriptsize
\begin{tabular}{lc|c|c|c|c|c|c|c|c}
\toprule
& Models & DropPatch & PatchTST & SimMTM & Ti-MAE & TST & LaST & CoST & TS2Vec \\

& Metrics & MSE ~ MAE & MSE ~ MAE & MSE ~ MAE & MSE ~ MAE & MSE ~ MAE & MSE ~ MAE & MSE ~ MAE & MSE ~ MAE \\
\midrule
\multirow{5}{*}{\shortstack{ETTh1\\$\downarrow$\\ETTm1}} 
& 96 & 0.290 ~ 0.345 & 0.289 ~ 0.344 & 0.290 ~ 0.348 & 0.667 ~ 0.521 & 0.425 ~ 0.381 & 0.295 ~ 0.387 & \textbf{0.248 ~ 0.332} & 0.605 ~ 0.561 \\
& 192 & 0.337 ~ 0.376 & 0.336 ~ 0.375 & \textbf{0.327 ~ 0.372} & 0.561 ~ 0.479 & 0.495 ~ 0.478 & 0.335 ~ 0.379 & 0.336 ~ 0.391 & 0.615 ~ 0.561 \\ 
& 336 & 0.366 ~ 0.395 & 0.365 ~ 0.395 & \textbf{0.357 ~ 0.392} & 0.690 ~ 0.533 & 0.456 ~ 0.441 & 0.379 ~ 0.363 & 0.381 ~ 0.421 & 0.763 ~ 0.677 \\ 
& 720 & 0.415 ~ 0.428 & 0.417 ~ 0.431 & \textbf{0.409 ~ 0.423} & 0.744 ~ 0.583 & 0.554 ~ 0.477 & 0.403 ~ 0.431 & 0.469 ~ 0.482 & 0.805 ~ 0.664 \\
& AVG & 0.352 ~ 0.386 & 0.352 ~ 0.386 & \textbf{0.346 ~ 0.384} & 0.666 ~ 0.529 & 0.482 ~ 0.444 & 0.353 ~ 0.390 & 0.359 ~ 0.407 & 0.697 ~ 0.616 \\
\midrule 
\multirow{5}{*}{\shortstack{ETTh2\\$\downarrow$\\ETTm1}} 
& 96 & 0.290 ~ 0.346 & 0.294 ~ 0.348 & 0.322 ~ 0.347 & 0.658 ~ 0.505 & 0.449 ~ 0.343 & 0.314 ~ 0.396 & \textbf{0.253 ~ 0.342} & 0.466 ~ 0.480 \\
& 192 & \textbf{0.344 ~ 0.379} & 0.345 ~ 0.379 & 0.332 ~ 0.372 & 0.594 ~ 0.511 & 0.477 ~ 0.407 & 0.587 ~ 0.545 & 0.367 ~ 0.392 & 0.557 ~ 0.532 \\ 
& 336 & 0.374 ~ 0.398 & \textbf{0.373} ~ 0.400 & 0.394 ~ \textbf{0.391} & 0.732 ~ 0.532 & 0.407 ~ 0.519 & 0.631 ~ 0.584 & 0.388 ~ 0.431 & 0.646 ~ 0.576 \\ 
& 720 & 0.435 ~ 0.437 & 0.444 ~ 0.437 & \textbf{0.411 ~ 0.424} & 0.768 ~ 0.592 & 0.557 ~ 0.523 & 0.368 ~ 0.429 & 0.498 ~ 0.488 & 0.752 ~ 0.638 \\
& AVG & \textbf{0.361} ~ 0.390 & 0.364 ~ 0.391 & 0.365 ~ \textbf{0.384} & 0.688 ~ 0.535 & 0.472 ~ 0.448 & 0.475 ~ 0.489 & 0.377 ~ 0.413 & 0.606 ~ 0.556 \\
\midrule 
\multirow{5}{*}{\shortstack{ETTm2\\$\downarrow$\\ETTm1}} 
& 96 & 0.285 ~ 0.339 & 0.289 ~ 0.347 & 0.297 ~ 0.348 & 0.647 ~ 0.497 & 0.471 ~ 0.422 & 0.304 ~ 0.388 & \textbf{0.239 ~ 0.331} & 0.586 ~ 0.515 \\
& 192 & \textbf{0.329 ~ 0.373} & 0.333 ~ 0.377 & 0.332 ~ 0.370 & 0.597 ~ 0.508 & 0.495 ~ 0.442 & 0.429 ~ 0.494 & 0.339 ~ 0.371 & 0.624 ~ 0.562 \\ 
& 336 & \textbf{0.356 ~ 0.392} & 0.363 ~ 0.398 & 0.364 ~ 0.393 & 0.700 ~ 0.525 & 0.455 ~ 0.424 & 0.499 ~ 0.523 & 0.371 ~ 0.421 & 1.035 ~ 0.806 \\ 
& 720 & \textbf{0.402 ~ 0.422} & 0.427 ~ 0.437 & 0.410 ~ 0.421 & 0.786 ~ 0.596 & 0.498 ~ 0.532 & 0.422 ~ 0.450 & 0.467 ~ 0.481 & 0.780 ~ 0.669 \\
& AVG & \textbf{0.343 ~ 0.382} & 0.353 ~ 0.390 & 0.351 ~ 0.383 & 0.682 ~ 0.531 & 0.480 ~ 0.455 & 0.414 ~ 0.464 & 0.354 ~ 0.401 & 0.756 ~ 0.638 \\
\midrule 
\multirow{5}{*}{\shortstack{Weather\\$\downarrow$\\ETTm1}} 
& 96 & \textbf{0.290 ~ 0.345} & 0.295 ~ 0.349 & 0.304 ~ 0.354 & - & - & - & - & - \\
& 192 & \textbf{0.331 ~ 0.373} & 0.343 ~ 0.378 & 0.338 ~ 0.375 & - & - & - & - & - \\ 
& 336 & \textbf{0.360 ~ 0.395} & 0.374 ~ 0.400 & 0.371 ~ 0.397 & - & - & - & - & - \\ 
& 720 & \textbf{0.411 ~ 0.427} & 0.424 ~ 0.434 & 0.417 ~ 0.426 & - & - & - & - & - \\
& AVG & \textbf{0.348 ~ 0.385} & 0.359 ~ 0.390 & 0.358 ~ 0.388 & - & - & - & - & - \\
\bottomrule
\end{tabular}
\end{table*}

\subsection{Full Results of Cold Start}
\label{sec:full_results_cold}
The full results under cold start scenario is presented in Table \ref{tab:cold_start_big}. 

\begin{table}[h]
\setlength{\tabcolsep}{2.5pt}
\caption{Full results under cold start scenario. Forecasting steps $T \in \{96, 192, 336, 720\}$. Models are pre-trained and fine-tuned in the same dataset.}
\label{tab:cold_start_big}
\centering
\scriptsize
\begin{tabular}{cc|c|c}
\toprule
&Models & DropPatch & PatchTST \\

&Metrics & MSE ~ MAE & MSE ~ MAE  \\
\midrule
\multirow{5}{*}{Exchange} 
& 96 & \textbf{0.081 ~ 0.199} & 0.083 ~ 0.202   \\
& 192 &\textbf{0.168 ~ 0.292} &	0.174 ~	0.297   \\ 
& 336 &\textbf{0.321 ~ 0.410} &	0.338 ~	0.419  \\ 
& 720 &\textbf{0.821 ~ 0.681} &	0.837 ~	0.689   \\
& AVG &\textbf{0.348 ~ 0.396} &	0.358 ~	0.402   \\

\midrule 

\multirow{5}{*}{PEMS03} 
& 12 & \textbf{0.081 ~ 0.194} &	0.082 ~	\textbf{0.194}   \\
& 24 &\textbf{0.127} ~ 0.240 &	\textbf{0.127 ~	0.237}   \\ 
& 48 &\textbf{0.219 ~ 0.317} &	0.227 ~	0.322  \\ 
& 96 &\textbf{0.363 ~ 0.419} &	0.383 ~	0.430   \\
& AVG &\textbf{0.198 ~ 0.293} &	0.205 ~	0.296   \\

\midrule 

\multirow{5}{*}{PEMS04} 
& 12 & \textbf{0.105} ~ 0.217 &	\textbf{0.105 ~	0.216}   \\
& 24 &0.165 ~ 0.274 &	\textbf{0.164 ~	0.271}   \\ 
& 48 &\textbf{0.291 ~ 0.371} &	0.300 ~	0.375  \\ 
& 96 &\textbf{0.493 ~ 0.495} &	0.522 ~	0.508   \\
& AVG &\textbf{0.264 ~ 0.339} &	0.273 ~	0.343   \\

\midrule 

\multirow{5}{*}{PEMS07} 
& 12 & \textbf{0.079 ~ 0.188} &	0.094 ~	0.220   \\
& 24 &\textbf{0.128 ~ 0.238} &	0.148 ~	0.280   \\ 
& 48 &\textbf{0.231 ~ 0.320} &	0.246 ~	0.349  \\ 
& 96 &\textbf{0.386 ~ 0.426} &	0.388 ~	0.442   \\
& AVG &\textbf{0.206 ~ 0.293} &	0.219 ~	0.323   \\

\midrule 

\multirow{5}{*}{PEMS08} 
& 12 & \textbf{0.091 ~ 0.199} &	0.097 ~	0.209   \\
& 24 & \textbf{0.139 ~ 0.243} &	0.141 ~	0.247   \\ 
& 48 & \textbf{0.247 ~ 0.327} &	0.253 ~	0.328  \\ 
& 96 & \textbf{0.423 ~ 0.430} &	0.439 ~	0.435   \\
& AVG & \textbf{0.225 ~ 0.300} &	0.233 ~	0.305   \\

\bottomrule
\end{tabular}
\end{table}

\label{sec:ablation}
\subsection{Parameter Sensitivity}
\label{sec:drop_mask_full}
\begin{figure*}[ht]
  \centering
  \includegraphics[width=\textwidth]{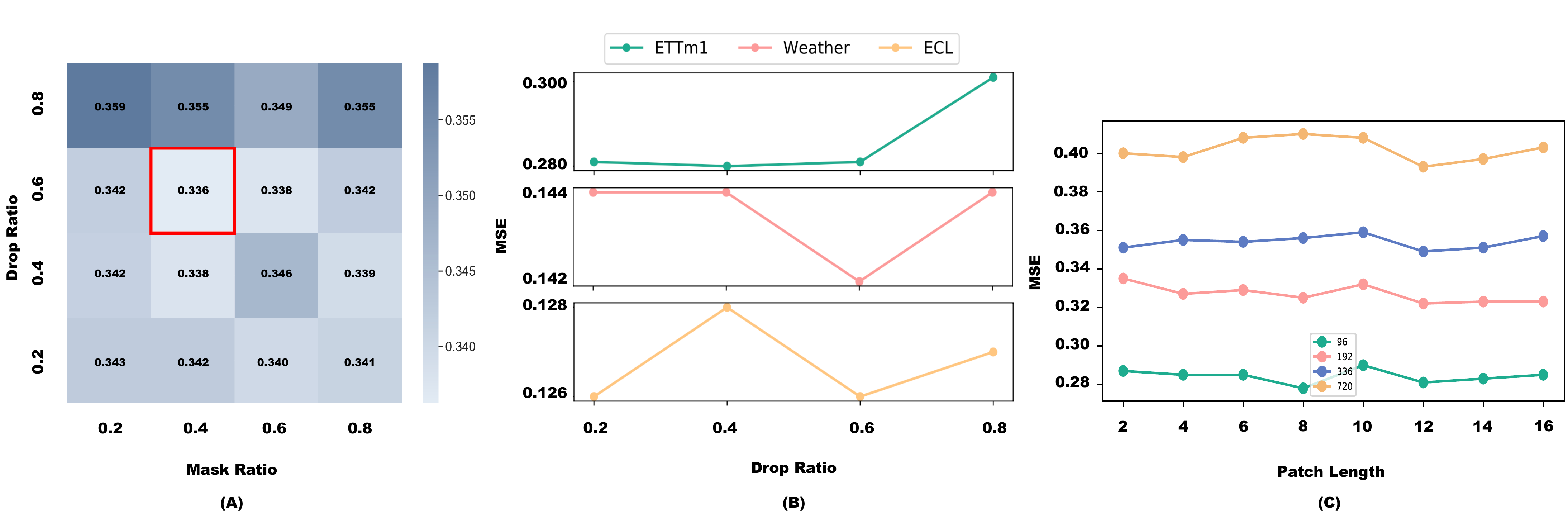}
  \caption{(A) MSE performance of DropPatch on ETTm1 dataset (in-domain setting) with all drop ratio and mask ratio. The values are averaged from different forecasting steps $T \in \{96, 192, 336, 720\}$. The lighter color denotes the better performance. (B) MSE performance of DropPatch adopting varying drop ratio $r$, mask ratio is fixed to be 0.4. (C) MSE performance of DropPatch on ETTm1 of forecasting steps $T \in \{96, 192, 336, 720\}$.
  \label{fig:model_analysis}}
\end{figure*}

Experiment results of sensitivity on drop and mask ratio for different forecasting steps are shown in Figure \ref{fig:drop_mask_full}, and the averaged results in \ref{fig:model_analysis} (A). We observed that the optimal mask ratio is 0.4, aligning with the official implementation of the PatchTST mask ratio. 

Consequently, we fixed the mask ratio at 0.4 and varied the drop ratio across different datasets. The result is displayed in Figure \ref{fig:model_analysis} (B). For each dataset, a drop ratio of 0.6 proves to be the reasonable choice, demonstrating the robustness and effectiveness of the DropPatch strategy.

We also study the effect of patch lengths on the forecasting performance of the ETTm1 dataset. We fix the lookback length $L_{pt}=L_{ft}=512$ and vary the patch length $ L_{P}\in \{2, 4, 6, 8, 10, 12, 14, 16\}$. In Figure \ref{fig:model_analysis} (C), we showcase the different forecasting results across various steps.

\section{Reproducibility}
The complete source code of DropPatch is available at https://github.com/qityy/DropPatch.

\section{Limitations and Future Works}
\label{sec:limitation}
The limitations and future works of this study mainly include two aspects:
\begin{enumerate}
    \item In this paper, we only discussed the strategy of randomly performing patch dropping and did not use other dropping strategies. In future work, we will explore different dropping strategies, such as those based on kernel density functions, binomial distributions, and clustering methods, to investigate their similarities and differences;
    \item Our synthesized dataset is still insufficient and not adequate to serve as a basis for training a foundation model. In future work, we will further seek stronger computational power and more data to advance towards a foundation model for time series.

\section{Potential Impact}
Our work is rooted in the field of time series modeling. In this specific field, time series exhibit a relatively low signal-to-noise ratio, and this characteristic directly leads to the rank collapse phenomenon being particularly prominent. It is worth noting that the rank collapse phenomenon is not exclusive to time series modeling and also exists widely in many other Transformer-based fields. From a theoretical perspective, the DropPatch we proposed has unique advantages. It can effectively slow down the rate of rank collapse in any field where representation learning is carried out with the help of Transformer. Based on theoretical grounds, DropPatch is, in principle, applicable to various different types of representation learning tasks. We sincerely hope that DropPatch and its derivative variants can not only effectively solve more complicated problems but also bring us richer and more diverse inspirations as well as deeper and more comprehensive thoughts in the process. 
\end{enumerate}

\end{document}